\documentclass{article}

\PassOptionsToPackage{numbers, compress}{natbib}

\usepackage{graphicx}
\usepackage{grffile}  
\usepackage{listings}
\usepackage[dblblindworkshop, final]{neurips_2026}

\workshoptitle{Responsible Communication of Machine Learning Research in Biomedicine}
\makeatletter
\renewcommand{\@noticestring}{Accepted at the RCMLR workshop (NeurIPS 2026)}
\makeatother

\usepackage[utf8]{inputenc} 
\usepackage[T1]{fontenc}    
\usepackage{hyperref}       
\usepackage{url}            
\usepackage{booktabs}       
\usepackage{amsfonts}       
\usepackage{nicefrac}       
\usepackage{microtype}      
\usepackage{xcolor}         
\usepackage{amsmath}
\usepackage{caption}
\usepackage{subcaption}
\usepackage{tabularx}
\usepackage{arydshln}

\usepackage[scr=boondox]{mathalpha}

\definecolor{new_color}{RGB}{200,230,255}

\title{DynSHAP: Towards Explainable Dynamic Survival Analysis}

\author{
  Nastasya Anokhina \\
  University of Cambridge\\
  \texttt{na634@cam.ac.uk} \\
  \And
  Jonas J\"ur\ss \\
  University of Cambridge\\
  \texttt{jj570@cl.cam.ac.uk} \\
  \And
  Pietro Li\`o \\
  University of Cambridge\\
  \texttt{pl219@cl.cam.ac.uk} \\
}

\begin{document}

\maketitle

\begin{abstract}
  Deep learning models for dynamic survival analysis (DSA) achieve strong predictive performance by incorporating longitudinal patient data, but their black box nature limits clinical trust and adoption. Existing explainability methods cannot handle longitudinal, irregular inputs and functional survival outputs simultaneously, which limits their usability in DSA. We propose \textit{DynSHAP}, a SHAP framework suited specifically for dynamic survival analysis. It extends common marginal SHAP estimators to this setting by treating time--feature pairs as players in the Shapley game. We further introduce \textit{Temporal DynSHAP}, which learns linear dependencies in features over time and uses conditional sampling to address them in explanations. When applied to synthetic data with known ground-truth attributions, \textit{Temporal DynSHAP} recovers temporally dependent features more accurately than marginal estimators for a given state-of-the-art model. Applied to two real-world clinical datasets and two DSA architectures, \textit{DynSHAP} produces attributions faithful to model learning, allowing medical experts to see which patient information drove the prediction and \textit{when}.

\end{abstract}

\section{Introduction}
\label{introduction}
\textit{Time-to-event} or \textit{Survival Analysis} is a statistical approach used to model the time until a certain event occurs. It has found important applications in healthcare to predict recovery, disease recurrence and the efficacy of treatments \citep{Lira2020-zx, fundamentals-and-applications-of-surv}. \textit{Dynamic Survival Analysis (DSA)} has expanded the method to incorporate longitudinal data in survival prediction. The applicability of DSA to the medical domain has been largely driven by the development of high-performing deep learning models such as Dynamic-DeepHit \citep{DDH} and DySurv \citep{mesinovic2024dysurvdynamicdeeplearning}.

While deep learning models demonstrate high performance in the field of survival analysis, the reasons for their predictions remain ambiguous. Practitioners cannot use DSA predictions confidently without knowing the clinical patterns underlying them. \textit{Explainable AI (XAI)} aims to make model decisions more transparent. In particular, post-hoc XAI methods analyse and interpret the decision-making process of a trained `black box' model after it has made predictions, providing insight into which features influenced the decision the most. One of the most well-known and applied frameworks for local post-hoc explainability is \textit{SHAP (SHapley Additive exPlanations)} \cite{lundberg2017unifiedapproachinterpretingmodel}. Recent studies have attempted to extend SHAP to time series data \citep{timeshap, nayebi2023windowshapefficientframeworkexplaining} as well as static survival analysis \citep{survshap}. 

However, none of these methods are applicable to \textit{dynamic} survival analysis. This limits applicability in the medical domain, where data is naturally presented as a sequence of irregular visits and disease progression is inherently a dynamic trajectory. Recent research \citep{Nguyen2023-xh, Ponce-Bobadilla2024-ii} in the clinical field reported that existing SHAP implementations did not account for temporal dependencies, which made them inapplicable to their longitudinal deep learning models. This further emphasises the existing gap between explainability and dynamic survival models.

To address this limitation, we propose DynSHAP, a SHAP framework that discretises the time domain into time--feature pairs and extends established marginal SHAP estimators to jointly handle longitudinal input and functional survival output. We further introduce a temporally-aware estimator that conditions on learned feature dependencies across time rather than sampling out-of-coalition values independently. In addition to estimating the importance of every feature, DynSHAP answers the question: \textit{at around what time did this feature most influence the predicted survival?}

The main contributions are summarised below:

\begin{enumerate}
    \item We propose a novel framework, DynSHAP, which extends SHAP to dynamic survival analysis by discretising the time domain and treating time--feature pairs as Shapley players. 
    \item We develop Temporal DynSHAP, which handles linear feature dependencies across time, and demonstrate its improved effectiveness in recovering ground truth.
    \item We provide an easy-to-use Python package for DynSHAP to facilitate usage by clinical practitioners. We apply the tool to real-world medical settings and showcase it recovers attributions consistent with established clinical findings.\footnote{The code is publicly available at \url{https://github.com/tasyaa04/dynShap}.}
\end{enumerate}

\lstset{
    backgroundcolor=\color{new_color},
    basicstyle=\ttfamily\small,
    language=Python,
    columns=fullflexible,
    keepspaces=true,
    frame=none,
    xleftmargin=0pt,
}
\begin{lstlisting}
explainer = DynamicExplainer(...)
explainer.fit(model, x_train, t_train, horizons)
explainer.plot()
\end{lstlisting}

\begin{figure} [htbp]
    \centering
    \includegraphics[width=\linewidth]{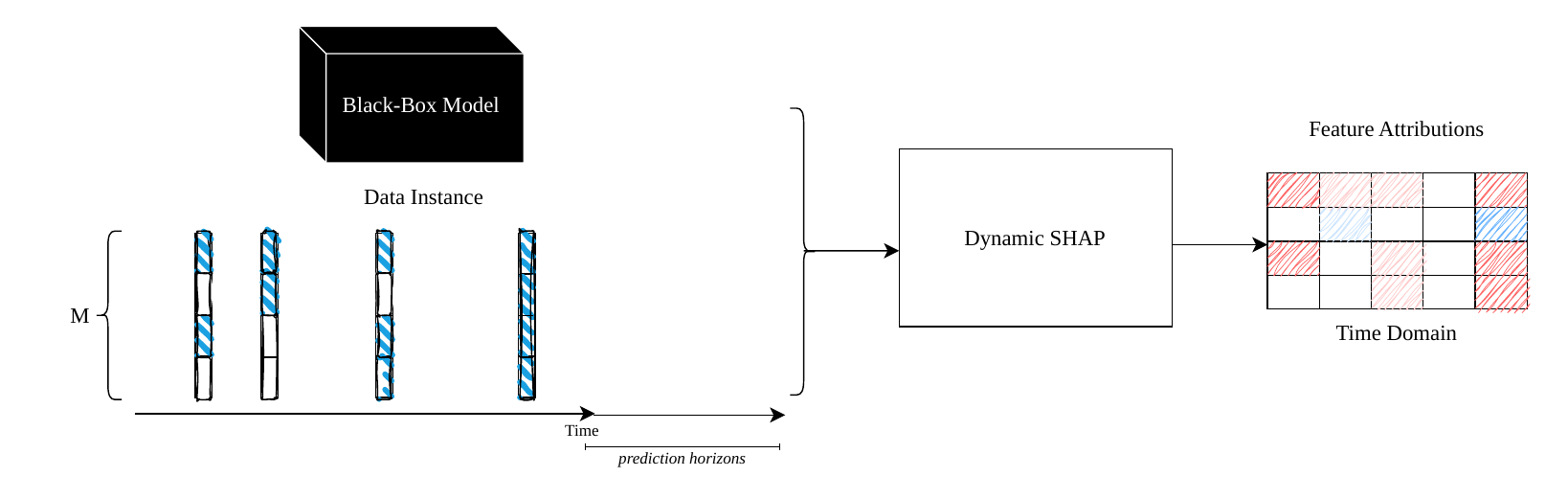}
    \caption{\textbf{High-level overview of how DynSHAP operates.} It takes a trained DSA model and a data instance to explain along with the prediction horizons and outputs local explanations for the data instance: time--feature attributions. The illustrated heatmap shows time--feature pairs that add and take away from the survival prediction in \textcolor{red}{red} and \textcolor{blue}{blue} respectively.}
    \label{fig:high-level-overview}
\end{figure}

\section{Related Work}
DSA models take irregular time series data as input and output survival functions. The input and output dimensions as well as black box modelling architectures present difficulties when applying existing SHAP methods to DSA models. These are concerned with usage convenience by non-technicians and the reliability of produced explanations, vital in the medical domain. 

\textit{SHAP (SHapley Additive exPlanations)} \citep{lundberg2017unifiedapproachinterpretingmodel} is a model-agnostic framework that assigns an importance value to each input feature considering its marginal contribution to the model’s output across all possible feature subsets. Traditional \textit{marginal} SHAP suffers from the correlation problem: simulating the absence of features by replacing them with sampled values from the background data can generate unrealistic data points. Various \textit{conditional} \citep{aas2020explainingindividualpredictionsfeatures} SHAP extensions have emerged to address this problem.

Techniques such as segmentation \citep{serramazza2025empiricalevaluationfactorsaffecting} have enabled the application of SHAP to time series data by treating time segments as Shapley players, without attributing importance to the individual features within the segments. Landmarking \citep{landmarking} followed by the application of SHAP to landmarks is a common approach in DSA as it allows the importance of features within time segments to be reported. However, this approach fails to account for the features' evolution over time and its usage by the model. More recent work has sought to address this limitation by extending SHAP specifically to time series. WindowSHAP \citep{nayebi2023windowshapefficientframeworkexplaining} and TimeSHAP \citep{timeshap} handle longitudinal input via window-grouping. Other approaches, such as SurvSHAP(t) \citep{survshap}, are designed to handle survival models with functional outputs.

However, to the best of our knowledge, neither method jointly handles irregular time series data and survival output conventions. Furthermore, existing approaches for estimating SHAP values in time series rely on marginal SHAP sampling, which can generate unrealistic data instances during the attribution process. These unrealistic instances may lead to unreliable model predictions, which are  propagated into the SHAP estimates. More importantly, such approaches may fail to capture the model's internal representation of temporal history and its dependence on previously observed information.  This may lead to recency biases by over-attributing the last timestamp. In healthcare, understanding both the clinical features and how their effects evolve over time is essential for producing timely and  meaningful predictions, particularly when early decisions can have a substantial impact on patient outcomes.

\section{SHAP Preliminaries}
\label{shap-background}

The concept of Shapley values originates from game theory and treats each model's feature as a player in a collaborative game. All players cooperate to produce the prediction of the model. The model's output is a reward to be distributed across players.

\paragraph{Definition} The \textit{Shapley value} of a feature value is its contribution to the model's prediction, weighted and summed over all possible feature value combinations:
\begin{equation}
\label{eq:exact_shap}
\phi_i(v) = \sum_{S \subseteq F \setminus \{i\}} \frac{|S|! (|F| - |S| - 1)!}{|F|!} (v(S \cup \{i\}) - v(S))
\end{equation}
where $\phi_i$ represents the contribution of the feature $i$ to the prediction, $F$ is the set of all features, $S$ is a subset of the features excluding $i$ and the value function $v_S$ is the payout of $S$ (the total reward the players in subset $S$ add).

\textit{Shapley Additive exPlanation (SHAP)} efficiently estimates Shapley values by looking at how a model's output deviates from its mean, one feature at a time.

\paragraph{Definition} Let $F$ be the set of all features; $f(X)$:  the prediction given a data vector $x^{(j)}$ of the model trained on $F$; $S$: a coalition, $S\subseteq F$; $C$: features not in coalition $S$, $C=F \setminus S$.
The \textit{value function} $v_{f,x^{(j)}}(S)$ is defined \cite{Molnar2022SHAP} as:
\begin{equation}
\label{eq:marginal_vf}
    v_{f,x^{(j)}}(S)=\int f\left( x_S^{(j)} \cup X_C \right) \mathrm{d}\mathbb{P}_{X_C}-\mathbb{E}[f(X)]
\end{equation}

The meaning of the value function is how much the features in the coalition $S$ add to the model's output. We can then calculate the SHAP value for each feature using Equation \ref{eq:exact_shap} and substituting $v_{f,x^{(j)}}(S)$ for $v(S)$.

There are $2^{|F|}$ possible coalitions of the feature values, which makes the computation of exact Shapley values intractable for most real-world applications.  The exponential complexity of this task and the scale of feature spaces have given rise to numerous estimation methods. The value function introduced in Equation \ref{eq:marginal_vf} is for \textit{marginal} or \textit{interventional} estimators. The marginal approach
measures a feature’s attribution to the model’s prediction by intervening on it while ignoring
correlations with other features. \textit{Conditional} SHAP addresses this issue by changing the value function \cite{Molnar2022SHAP} to:
\begin{equation}
\label{eq:conditional_vf}
    v_{f,X}(S)=\int f\left(x_S^{(i)} \cup X_C\right)d\mathbb{P}_{X_C\mid X_S=x_S^{(i)}}-\mathbb{E}[f\left(X\right)]
\end{equation}

\begin{figure}[htb]
    \centering
    \includegraphics[width=\linewidth]{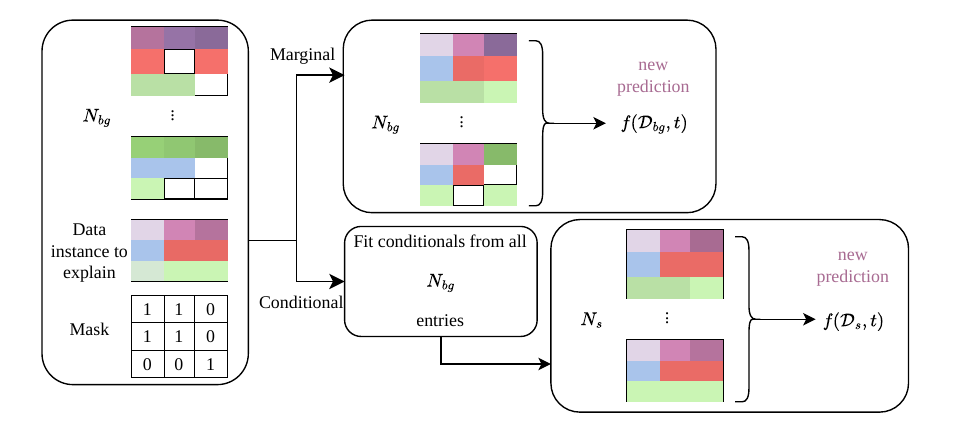}
    \caption{\textbf{The difference between marginal and conditional SHAP sampling for one coalition mask.} One matrix cell indicates a time--feature pair. All background data cells that are `present' (mask cell value = 1) are substituted with the cell value from the data instance being explained. In the marginal case, all `missing' cells (mask cell value = 0) are left as before. In the conditional case, all background data entries are used to fit the conditionals. Then, the cells that are `missing' are substituted with the values predicted from the conditionals given new masked values. We see that for the feature in the first row, conditional sampling produces more realistic feature trajectories that are then used in model prediction.}
    \label{fig:marginal-vs-conditional-ill}
\end{figure}

Crucially, marginal SHAP aims to answer the question ``\textit{what would the model predict if we intervened on feature $X$?}'', whereas conditional SHAP answers ``\textit{what would the model predict given a realistic sample with the intervened feature value $X$?}''. There is an ongoing debate \cite{janzing2019featurerelevancequantificationexplainable} on which approach is preferable. Patient trajectories likely have many correlations across time, so this problem is worth exploring. The differences in sampling and masking strategies are illustrated in Figure~\ref{fig:marginal-vs-conditional-ill}.

\section{Methods}
\label{methods}

\subsection{Problem Setup}

\begin{figure}[htbp]
    \centering
    \includegraphics[width=\linewidth]{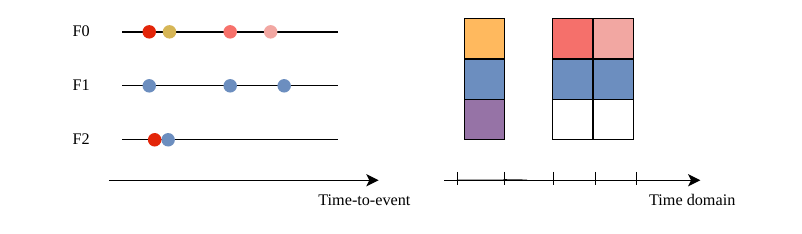}
    \caption{\textbf{The binning process of every data instance being explained.} Feature values that fall in a single time bin are aggregated to yield distinct time--feature pairs. Each time--feature pair is the player in the SHAP estimation game.}
    \label{fig:binning-illustration}
\end{figure}

A patient trajectory $x_i \in \mathbb{R}^{T_i \times M}$ consists of $M$ features measured at irregular times, with $T_i$ being different for each patient $i$. We discretise the time domain into $K$ bins of resolution $r$: $K = \lceil \max(T^*)/r \rceil$, and define each feature $f_m$ at bin $t_k$ as the mean of all measurements falling in that bin (Figure~\ref{fig:binning-illustration}). Each time--feature pair $(t_k, f_m)$ is treated as an independent player, giving a feature space of size $P = K \times M$. This preserves temporal structure and lets attribution vary across time.

When handling model output, we avoid collapsing the survival function to a single scalar. We follow SurvSHAP(t) \cite{survshap} and compute separate SHAP estimates at each prediction horizon.

\subsection{Marginal Estimators}
\label{dynamic-sampling-shap}

\paragraph{Sampling DynSHAP}
\label{monte-carlo}
We extend the Monte Carlo Sampling estimator \citep{CASTRO20091726, mitchell2022sampling} to time--feature pairs. For each of $B$ random permutations, we process the pairs in order. At position $k$ in the permutation, the marginal contribution of pair $\pi(k)$ is:
\begin{equation}
\label{eq:delta-sampling-shap}
    \Delta_{\pi(k)} = f\!\left(\{\pi(1), \ldots, \pi(k)\}\right) - f\!\left(\{\pi(1), \ldots, \pi(k-1)\}\right)
\end{equation}

The SHAP value for pair $i$ is the average of its marginal contributions across all permutations:
\begin{equation}
\label{eq:perm-dyn}
    \hat\phi_i^{\text{perm}} = \frac{1}{B} \sum_{b=1}^{B} \Delta_i^{(b)}
\end{equation}

\paragraph{Kernel DynSHAP}
\label{dynamic-kernel-shap}
We sample masks uniformly at random from $\{0,1\}^P$, each assigned a weight according to the Shapley kernel~\citep{lundberg2017unifiedapproachinterpretingmodel}:
\begin{equation}
    \pi(\mathbf{z}) = \frac{P - 1}{\binom{P}{|\mathbf{z}|} \cdot |\mathbf{z}| \cdot (P - |\mathbf{z}|)},
\end{equation}
where $|\mathbf{z}| = \sum_i z_i$ is the coalition size.

For each mask $\mathbf{z}$, present pairs ($z_{(t_k,f_m)}=1$) take the target patient's value; absent pairs take background values at the corresponding visits. Averaging over background patients yields the marginal value function:
\begin{equation}
\label{eq:marginal-vf-dynamic}
    v(\mathbf{z}) = \frac{1}{N_{\text{bg}}} \sum_{j=1}^{N_{\text{bg}}} f\!\left(\mathbf{x}^{(j)}_{\mathbf{z}}, \mathbf{t}_{out}\right)
\end{equation}
where $\mathbf{x}^{(j)}_{\mathbf{z}}$ is the hybrid observation for background patient $j$ and $f(\cdot, \mathbf{t}_{out})$ is the predicted survival function at horizon $\mathbf{t}_{out}$.

Given the matrix of coalition masks $\mathbf{Z} \in \{0,1\}^{N \times P}$ (with $N$ sampled coalitions), the diagonal weight matrix $\mathbf{W} = \text{diag}(\pi(\mathbf{z}_1), \ldots, \pi(\mathbf{z}_N))$, and the vector of value function evaluations $\mathbf{y} \in \mathbb{R}^{N \times H}$, Shapley values are recovered by weighted least squares:

\begin{equation}
\boldsymbol{\Phi} =
(\mathbf{Z}^{\top} \mathbf{W} \mathbf{Z})^{-1}
\mathbf{Z}^{\top} \mathbf{W} \mathbf{y}
\label{eq:kernel_solve}
\end{equation}

\subsection{Temporal DynSHAP}
\label{dynamic-temporal-shap}

We assume linear dependence between features over time. The justification for this choice is presented in Appendix \ref{appendix-temp-dep}. To address feature dependence over time, we adapt the value function (\ref{eq:marginal-vf-dynamic}) to account for the new conditional assumption. The new conditional value function is:
\begin{equation}
    v_{\text{cond}}(\mathbf{z}) = \frac{1}{N_s} \sum_{s=1}^{N_s} f\!\left( \tilde{\mathbf{x}}^{(s)}_{\mathbf{z}},\mathbf{t}_{out}\right)
\end{equation}
where each $\tilde{\mathbf{x}}^{(s)}_{\mathbf{z}}$ is constructed by keeping the target patient's values for in-coalition pairs and sampling from learned temporal conditionals for out-of-coalition pairs. The average over $N_s$ conditional draws approximates the conditional expectation. Figure~\ref{fig:marginal-vs-conditional-ill} illustrates how this approach differs.

We choose a linear Gaussian model as the distribution from which the out-of-coalition players are sampled. The features at $t = 0$ are sampled marginally from the background data. For subsequent time bins $t \geq 1$:
\begin{equation}
    x_f^{(t)} \approx \mathbf{w}_{t,f}^\top \mathbf{x}^{(t-1)} + b_{t,f} + \varepsilon, \quad \varepsilon \sim \mathcal{N}(0, \hat{\sigma}_{t,f}^2)
\end{equation}
fit with $\ell_2$ regularisation on the binned, forward/backward-filled background dataset.

\section{Experiments}
\label{experiments}

We trained and evaluated two DSA models Dynamic-DeepHit \citep{DDH} and DySurv \citep{mesinovic2024dysurvdynamicdeeplearning} on two real-world datasets and synthetic data. Both models exceed 0.7 concordance index on all datasets (Appendix \ref{appendix-model-training}), indicating strong predictive discrimination and making them suitable candidates for interpretation to detect which learned patterns result in satisfactory accuracy. We used the resolution $r=1$ as a parameter in generating all DynSHAP explanations. Future experiments could include sensitivity analysis by comparing explanations produced with different values of $r$.

\subsection{Temporal Dependence Recovery} 
\label{synth-exp}
Because real-world data usually has no defined ground truth, explainability tools are often verified on synthetic data, especially in the case of conditional SHAP \citep{survshap, ng2025causal}. To test whether Temporal DynSHAP recovers temporally dependent features better and compare DynSHAP with baselines, we generate a synthetic longitudinal survival dataset. The data is designed with four covariates with strong dependence over time to test that Temporal DynSHAP handles such features better. Further details on the synthetic dataset and ground truth definition are outlined in Appendix \ref{synthetic-data-generation}.

We evaluate all estimators and baselines using RMSE and MAE between their SHAP estimates and the ground truth. Table \ref{tab:gt-agreement} shows that Temporal DynSHAP recovers ground truth $\sim2\times$ better than its marginal counterparts when applied to DySurv, although this advantage does not hold for DDH. Generally, all methods perform substantially worse on DDH. This may suggest that DDH fails to learn the linear correlations between features on the given dataset. Future work could investigate this hypothesis by evaluating it on datasets with varied linear correlations across time.

\begin{table}[t]
\centering
\resizebox{\textwidth}{!}{%
\begin{tabular}{lcccc}
\toprule
& \multicolumn{2}{c}{DySurv} & \multicolumn{2}{c}{DDH} \\
\cmidrule(lr){2-3} \cmidrule(lr){4-5}
Estimator & RMSE & MAE & RMSE & MAE \\
\midrule
Temporal DynSHAP (ours) & $\mathbf{0.0032 \pm 0.0005}$ & $\mathbf{0.0005 \pm 0.0001}$ & $0.0273 \pm 0.0020$ & $0.0131 \pm 0.0015$ \\
Kernel DynSHAP (ours)   & $0.0064 \pm 0.0006$ & $0.0019 \pm 0.0002$ & $0.0135 \pm 0.0001$ & $0.0057 \pm 0.0000$ \\
Sampling DynSHAP (ours) & $0.0065 \pm 0.0006$ & $0.0018 \pm 0.0001$ & $\mathbf{0.0134 \pm 0.0001}$ & $\mathbf{0.0057 \pm 0.0000}$ \\
\hdashline
Random noise (control)  & $0.0077 \pm 0.0002$ & $0.0061 \pm 0.0002$ & $0.0265 \pm 0.0002$ & $0.0195 \pm 0.0001$ \\
Landmark SHAP           & $0.0269 \pm 0.0000$ & $0.0066 \pm 0.0000$ & $0.0322 \pm 0.0000$ & $0.0165 \pm 0.0000$ \\
\bottomrule
\end{tabular}%
}
\vspace{6pt}
\caption{\textbf{Agreement with ground-truth SHAP on the synthetic dataset.} Calculated with Root Mean Squared Error (RMSE) and Mean Absolute Error (MAE) between SHAP estimates and ground truth Shapley value, lower is better. Reported mean $\pm$ std over 5 seeds.}
\label{tab:gt-agreement}
\end{table}

\subsection{Practical Applications}
\label{qualitative}

The main motivation of DynSHAP is practical applications in the medical domain. Patient data are rarely presented as static events and timely attributions to clinical markers are necessary for early intervention when required. Additionally, technicians can use DynSHAP to infer model learning and change architectural decisions to build more reliable models. DynSHAP permits to obtain such explanations but their evaluation on real world data is challenging due to the lack of ground truth knowledge.

We test DynSHAP on two real world datasets. The first dataset is the longitudinal \texttt{PBC} dataset \cite{murtaughPrimaryBiliaryCirrhosis1994}, derived from the Mayo Clinic trial on primary biliary cirrhosis of the liver conducted between 1974 and 1984. It consists of 312 patients and 18 features of different types, including continuous, discrete and binary. The second is a dataset simulated by sampling from anonymised data of patients. It comprises 500 patients and 98 features after pre-processing. The features are mainly binary, indicating diagnoses and treatments prescribed. The features were filtered by their density in the dataset: only the features present in at least 10 MS cases recorded were kept. The event of interest is Multiple Sclerosis (MS) development. 

We conduct a faithfulness test to verify that the estimators' explanations are truthful to what the models learn, following the standard perturbation experiment \citep{samek-perturb}. Prior to deletion, the cells (time--feature pairs) are grouped into clusters based on their correlation to account for feature dependence in the data and make the results less susceptible to OOD data instances. Each cluster’s importance is the sum of its cells’ absolute attributions, averaged over the prediction horizons. Clusters are ranked by this score and removed in order, masking the corresponding cells with background data, following the standard SHAP-style marginal masking. As shown in Table \ref{tab:faithfulness}, our methods outperform baselines for the synthetic and PBC data, indicating greater faithfulness to model learning. The results are more noisy on the MS dataset, which could be explained by the sparsity of the data. The DDH results favour the Temporal estimator more, potentially indicating that DDH's predictions are more susceptible to OOD inputs. Notably, the marginal estimator (Kernel DynSHAP) performs better than the conditional one (Temporal DynSHAP) on the DySurv model, contrary to the results of the ground truth experiment. The faithfulness metrics directly favour marginal approaches, since they replace the masked features with background data, matching marginal sampling. Temporal DynSHAP assumes linear dependencies in all features over time and, hence, might create spurious correlations where there are none in the data, or fail to account for longer-range dependencies. Future work could explore other conditional distributions for the sampling process.

\begin{figure}[htbp]
    \centering
    \begin{subfigure}[b]{\textwidth}
        \centering
        \includegraphics{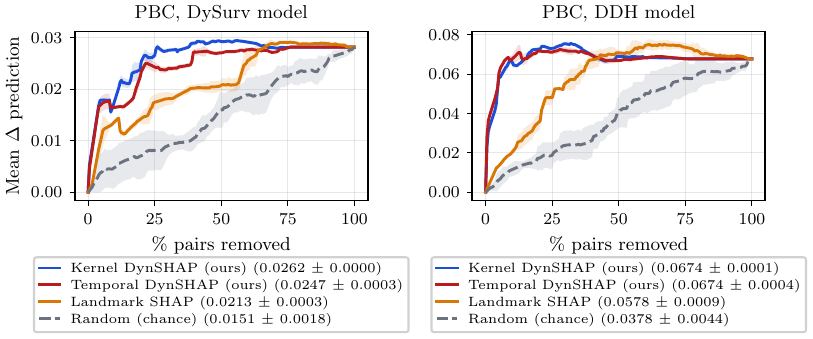}
        \label{fig:pbs-faith}
    \end{subfigure}

    \vspace{1em}
    
    \caption{\textbf{Most Important Feature (MIF) ablation curves comparing DynSHAP against baselines.} The plots evaluates feature ranking accuracy by measuring the mean absolute change in the model's prediction as top-ranked clusters of time--feature pairs are progressively removed. Aggregated across 5 patients, reported mean $\pm$ std across 5 seeds.}
    \label{fig:faithfulness}
\end{figure}

\begin{table}[t]
\centering
\resizebox{\textwidth}{!}{%
\begin{tabular}{lcccccc}
\toprule
& \multicolumn{2}{c}{PBC} & \multicolumn{2}{c}{MS} & \multicolumn{2}{c}{Synthetic} \\
\cmidrule(lr){2-3} \cmidrule(lr){4-5} \cmidrule(lr){6-7}
Estimator & DySurv & DDH & DySurv & DDH & DySurv & DDH \\
\midrule
Kernel DynSHAP (ours)   & $\mathbf{0.0262 \pm 0.0000}$ & $\mathbf{0.0674 \pm 0.0001}$ & $0.0094 \pm 0.0002$ & $0.0034 \pm 0.0000$ & $\mathbf{0.0213 \pm 0.0002}$ & $\mathbf{0.0357 \pm 0.0009}$ \\
Temporal DynSHAP (ours) & $0.0247 \pm 0.0003$ & $\mathbf{0.0674 \pm 0.0004}$ & $0.0079 \pm 0.0006$ & $\mathbf{0.0039 \pm 0.0001}$ & $\mathbf{0.0213 \pm 0.0003}$ & $0.0289 \pm 0.0026$ \\
\hdashline
Landmark SHAP           & $0.0213 \pm 0.0003$ & $0.0578 \pm 0.0009$ & $0.0068 \pm 0.0004$ & $0.0036 \pm 0.0002$ & $0.0206 \pm 0.0000$ & $0.0326 \pm 0.0000$ \\
Random (chance)         & $0.0151 \pm 0.0018$ & $0.0378 \pm 0.0044$ & $\mathbf{0.0097 \pm 0.0028}$ & $0.0029 \pm 0.0006$ & $0.0104 \pm 0.0047$ & $0.0219 \pm 0.0036$ \\
\bottomrule
\end{tabular}%
}
\vspace{6pt}
\caption{\textbf{Mean change in prediction across datasets and model types.} Calculated by clustering correlated features, ranking them by estimated importance and removing groups in order of most important first. Higher change in model's output signifies the explanations are more faithful to model learning. Reported mean $\pm$ std across 5 seeds.}
\label{tab:faithfulness}
\end{table}

\begin{figure}
     \centering
    \begin{subfigure}[b]{0.30\textwidth}
        \centering
        \includegraphics[width=\linewidth]{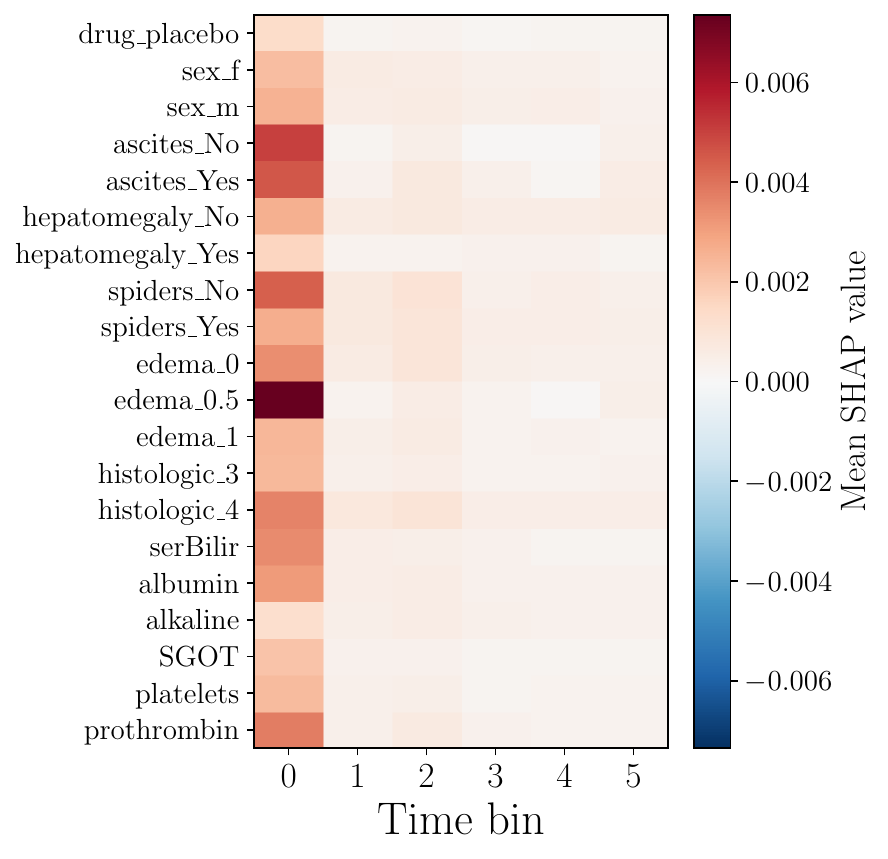}
        \caption{Kernel DynSHAP, DDH}
        \label{fig:kernel-ddh--pbc}
    \end{subfigure}
    \hfill
    \begin{subfigure}[b]{0.30\textwidth}
        \centering
        \includegraphics[width=\linewidth]{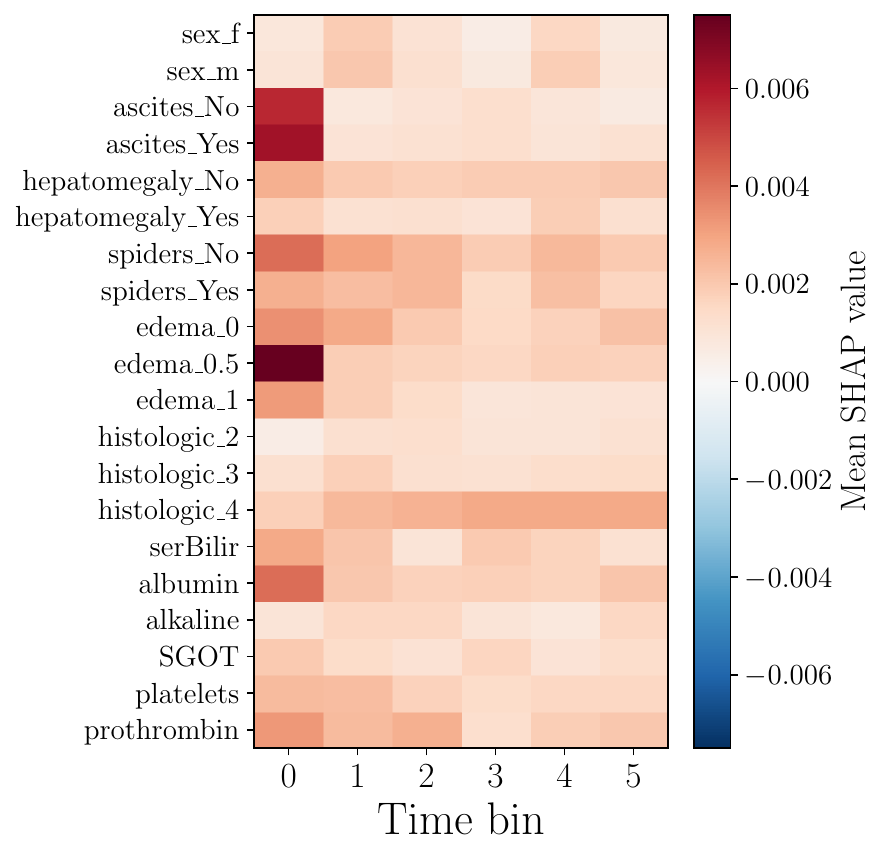}
        \caption{Temporal DynSHAP, DDH}
        \label{fig:temporal-ddh-pbc}
    \end{subfigure}
    \hfill
    \begin{subfigure}[b]{0.30\textwidth}
        \centering
        \includegraphics[width=\linewidth]{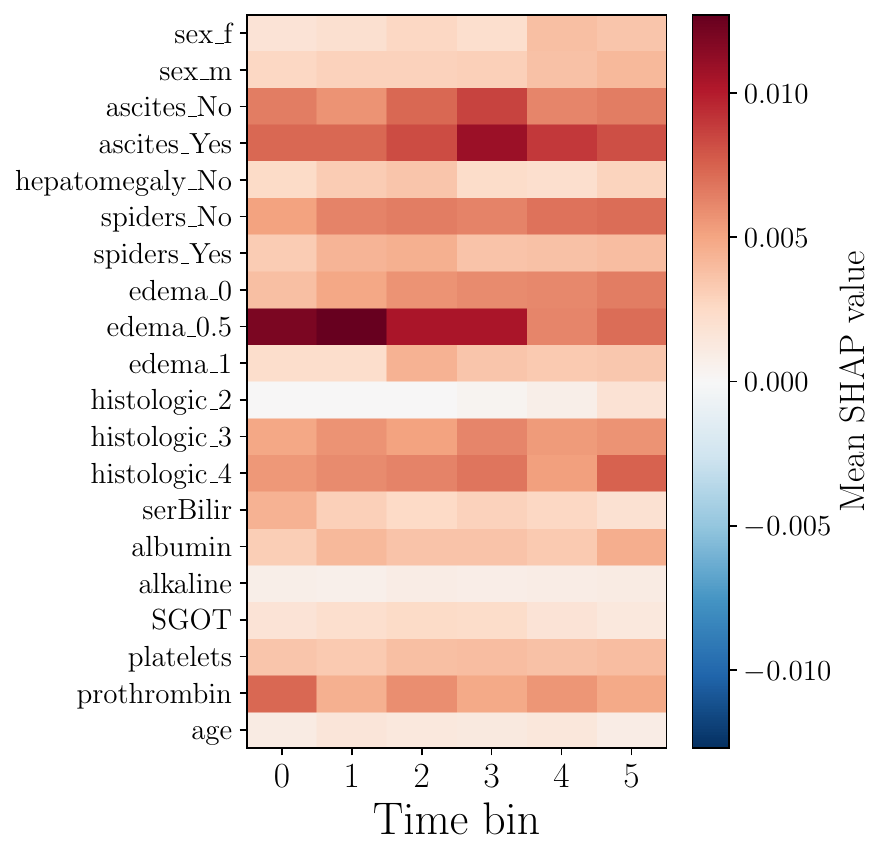}
        \caption{Landmark DynSHAP, DDH}
        \label{fig:landmark-ddh-pbc}
    \end{subfigure}

    \vspace{1em}

    \centering
    \begin{subfigure}[b]{0.30\textwidth}
        \centering
        \includegraphics[width=\linewidth]{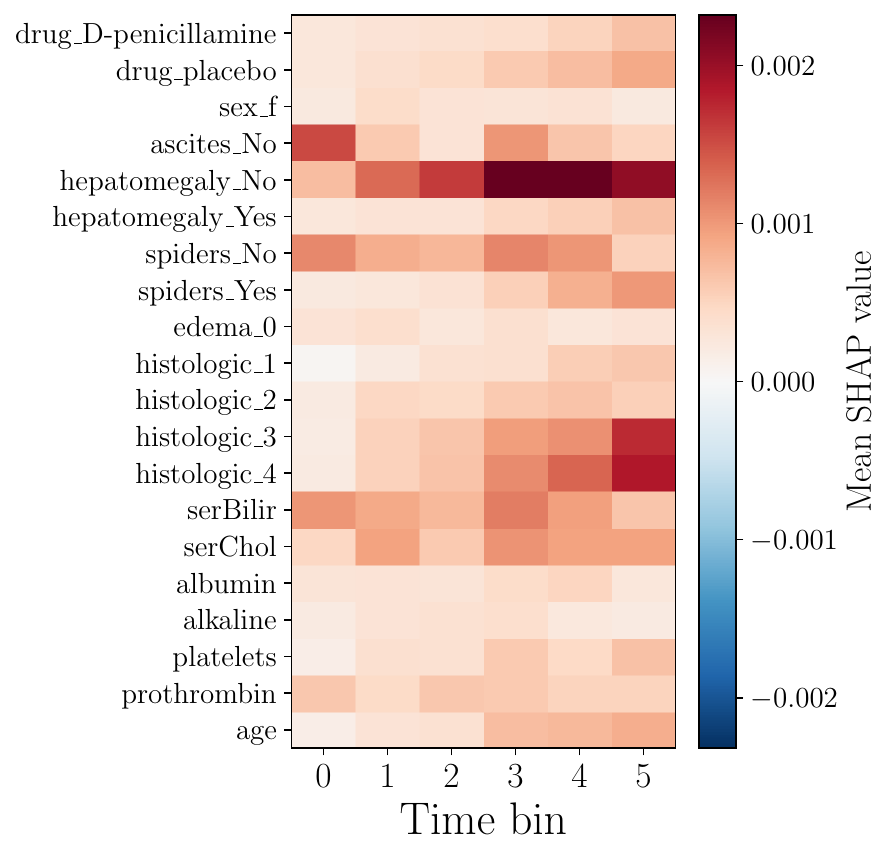}
        \caption{Kernel DynSHAP, DySurv}
        \label{fig:kernel-dysurv-pbc}
    \end{subfigure}
    \hfill
    \begin{subfigure}[b]{0.30\textwidth}
        \centering
        \includegraphics[width=\linewidth]{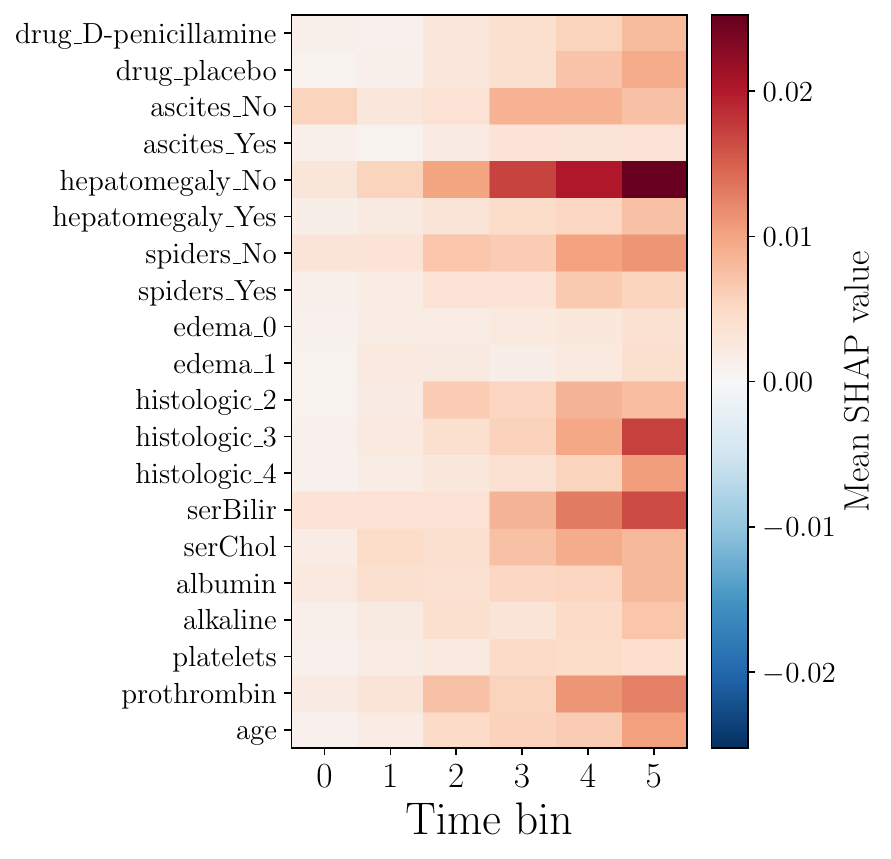}
        \caption{Temporal DynSHAP, DySurv}
        \label{fig:temporal-dysurv-pbc}
    \end{subfigure}
    \hfill
    \begin{subfigure}[b]{0.30\textwidth}
        \centering
        \includegraphics[width=\linewidth]{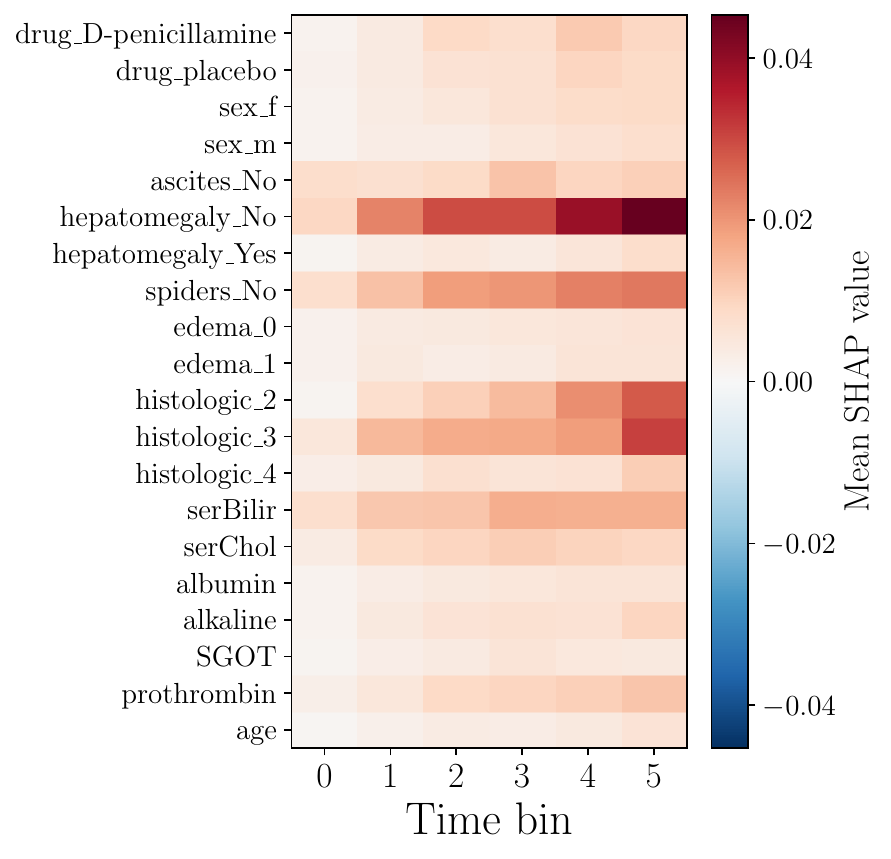}
        \caption{Landmark DynSHAP, DySurv}
        \label{fig:landmark-dysurv-pbc}
    \end{subfigure}
     \caption{\textbf{A comparison of SHAP estimators on DSA models trained on PBC data.} Demonstrated on absolute mean estimated SHAP values aggregated across a population of 62 patients from the full test dataset. Applied to two state-of-the-art deep learning DSA models.}
     \label{fig:pbc_pop}
 \end{figure}

 Figure \ref{fig:pbc_pop} demonstrates the difference between the estimations of learned patterns by both models trained on the PBC dataset. The estimators identify features aligned with established clinical literature. Critical continuous biomarkers, such as albumin and bilirubin, along with physical symptoms like edema and ascites, are correctly highlighted as primary drivers of the survival prediction by both Kernel and Temporal DynSHAP. The attribution of risk to Albumin and Prothrombin matches the risks identified in the Mayo model \cite{Dickson1989PrognosisIP}. The estimators' attribution to earlier timestamps differs: the marginal estimator collapses everything into the last timestamp for the DDH model, whereas the temporal one attributes more to historical data. The landmarking explanations seem plausible but the method does not use the patient's accumulated visit history when generating a prediction for a given landmark, and so cannot be a very faithful representation of model history learning. Additionally, it violates local accuracy (see Appendix \ref{appendix-local-accuracy-landmark}) and in some construction cases, missingness. These are two important SHAP properties re-defined for the DSA setting in Appendix \ref{appendix-shap-props}. Moreover, Table \ref{tab:faithfulness} demonstrates empirically that the landmarking approach is less faithful to the models' learning. Appendix \ref{appendix-ms-case-study} presents additional qualitative analysis for the patient from the MS dataset. Overall, our tool can be used to understand a `black box' model's learning better and check how reliable it truly is, which is exactly the purpose of XAI. 

\section{Conclusion}
\label{conclusion}
We demonstrate that existing techniques are only partially sufficient to explain dynamic survival analysis models. This is an important gap in healthcare where patient histories are recorded over time and can be used in disease and survival prediction. We propose the DynSHAP framework, which extends traditional SHAP methods to dynamic survival analysis and analyse its robustness and applicability to real world settings. Unlike traditional SHAP tools, DynSHAP handles longitudinal, irregular time data and functional model output. It further makes a step towards addressing feature dependence over time in survival explanations. The synthetic experiments show that Temporal DynSHAP recovers temporally dependent features better than marginal estimators for one of the state-of-the-art DSA models but not the other. Additionally, DynSHAP appears to be more faithful to model learning than the baselines for the majority of models and datasets. On real clinical datasets, all DynSHAP estimators produce plausible explanations supported by medical literature.

Future work should explore temporal conditionals beyond the linear-Gaussian assumption to better capture non-linear or longer-range dependencies between visits and improve both faithfulness and stability. A causally-informed value function, conditioning on an estimated or known causal graph over time--feature pairs rather than purely predictive dependence, is a further natural extension, particularly in settings where such structure can be estimated reliably. Validating DynSHAP on larger and more diverse clinical cohorts, and directly against clinician judgement, would strengthen evidence for its practical utility as a tool for auditing and trusting dynamic survival models in deployment. Finally, future experiments could investigate why certain estimators are better suited to specific model architectures, using a wider range of synthetic data.

\begin{ack}
We would like to thank Ali Manouchehrinia and Sonia Darvishi for the MS dataset and fruitful discussions as well as the reviewers of the RCMLR workshop (NeurIPS 2026) for their insightful feedback. Jonas Jürß is funded by Horizon Europe (WISDOM project Grant No. 101137154).
\end{ack}


\bibliographystyle{unsrtnat}
\bibliography{references}

\appendix

\section{On DynSHAP Properties}
\label{appendix-shap-props}
We redefine SHAP's three desirable properties \citep{lundberg2017unifiedapproachinterpretingmodel} for our problem setting:

Let:
\begin{itemize}
    \item $f(\mathbf{x},\mathbf{t}_{out}):(\chi, \mathbb{R}^H) \rightarrow \mathbb{R}^H$ denote the model's \texttt{predict(\texttt{data}, \texttt{timestamps})}.
    \item $\mathcal{P}=\{(t_i,f_i)\}_{i=1}^P$ be the set of time--feature pairs.
    \item $\phi_i(f,\mathbf{x}) \in \mathbb{R}^H$ be the vector of SHAP values of pair $i$ at each of $\mathbf{t}_{out}$ timestamps.
    \item $\mathbb{E}[f(X,\mathbf{t}_{out})] \in \mathbb{R}^H$ be the baseline mean survival curve over the background data.
\end{itemize}

We then formulate the three properties our estimators should have:

\paragraph{Theorem 1: Local Accuracy.} For any DSA model $f$ and features $X$, SHAP values must satisfy:
\begin{equation}
    \sum_{i \in P} \phi_i(f,\mathbf{x}) = f(\mathbf{x},\mathbf{t}_{out}) - \mathbb{E}[f(X,\mathbf{t}_{out})]
\end{equation}

This means that the predicted survival curve at instance $\mathbf{x}$ must equal the baseline survival curve plus the sum of each time--feature pair's contribution curve, so that local accuracy holds at each predicted point of survival.

\paragraph{Theorem 2: Missingness} If, for a given patient, the feature $\mathscr f$ is never recorded in the time bin $i$, the pair $(t_i,\mathscr f)$ gets an attribution of zero. Formally:
\begin{equation}
    \mathbf{x}_\texttt{binned}[t_i][\mathscr f] = \text{NaN} \implies \phi_{(t_i,\mathscr f)}(f,\mathbf{x})=\mathbf{0} \in \mathbb{R}^H
\end{equation}

\paragraph{Theorem 3: Consistency} Let $f_x(z',\mathbf{t}_{out})=f(h_x(z'),\mathbf{t}_{out})$, where $h_x$ maps simplified binary inputs $z' \in \{0,1\}^{P}$ to the original longitudinal inputs, and let $z' \setminus i$ denote setting the indicator for a specific time--feature pair $i \in \mathcal{P}$ to $0$. For any two DSA models $f$ and $f'$, if the marginal contribution of pair $i$ satisfies:
\begin{equation}
    f'_x(z',\mathbf{t}_{out})-f'_x(z'\setminus i,\mathbf{t}_{out}) \geq f_x(z',\mathbf{t}_{out})-f_x(z'\setminus i,\mathbf{t}_{out})
\end{equation}
for all possible subsets of time--feature pairs $z' \in \{0,1\}^{P}$, then the attribution for that pair must not decrease:
\begin{equation}
    \phi_i(f',\mathbf{x}) \geq \phi_i(f,\mathbf{x}),
\end{equation}
Note that the operators $\geq$ and $-$ are pointwise, so consistency is satisfied at every predicted timestamp.

The implemented estimators should adhere to all these properties. The marginal estimators are direct extensions of the original SHAP algorithms \cite{lundberg2017unifiedapproachinterpretingmodel}. As such, they theoretically inherit SHAP's desirable properties. Temporal DynSHAP is designed to inherit the functionality from Kernel DynSHAP to adhere to the properties as well.

\subsection{Violations by landmarking}
\label{appendix-local-accuracy-landmark}
\begin{figure}[htb]
    \centering
    \begin{subfigure}{0.48\textwidth}
        \centering
        \includegraphics[width=\linewidth]{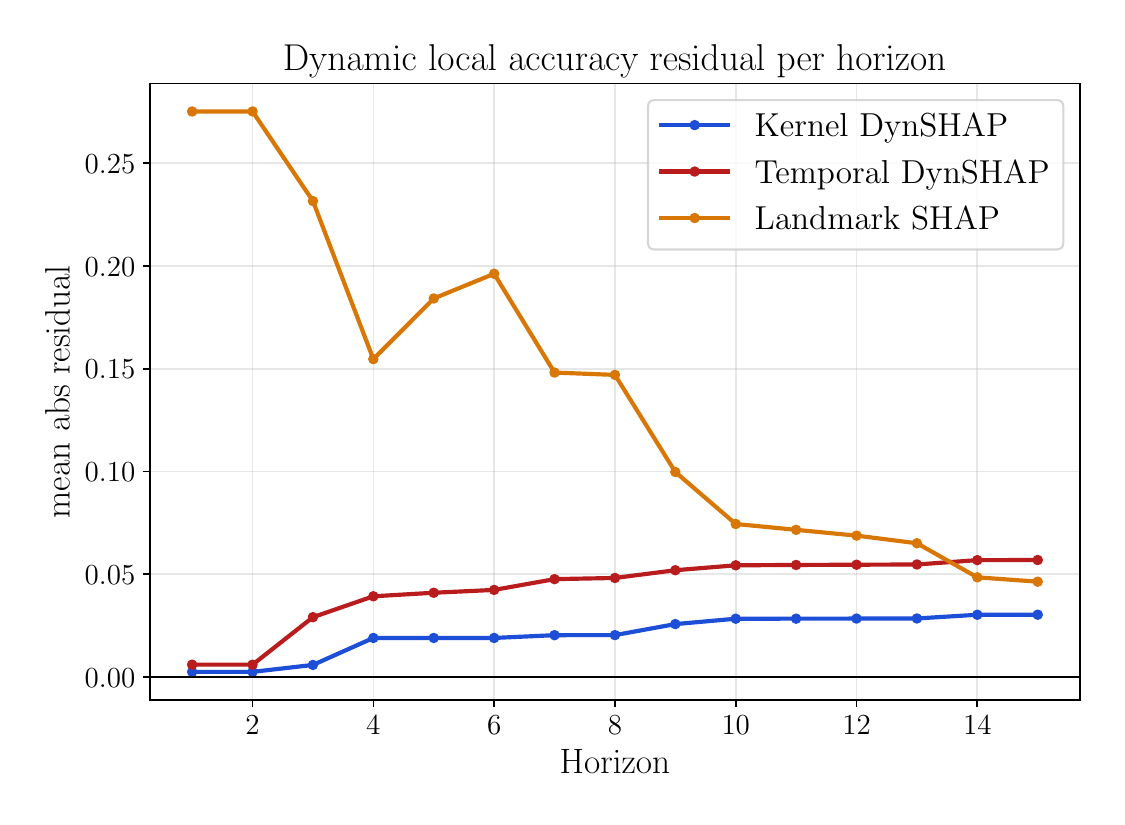}
        \caption{Local accuracy residuals, DySurv, MS.}
        \label{fig:local_acc_ms}
    \end{subfigure}
    \hfill 
    \begin{subfigure}{0.48\textwidth}
        \centering
        \includegraphics[width=\linewidth]{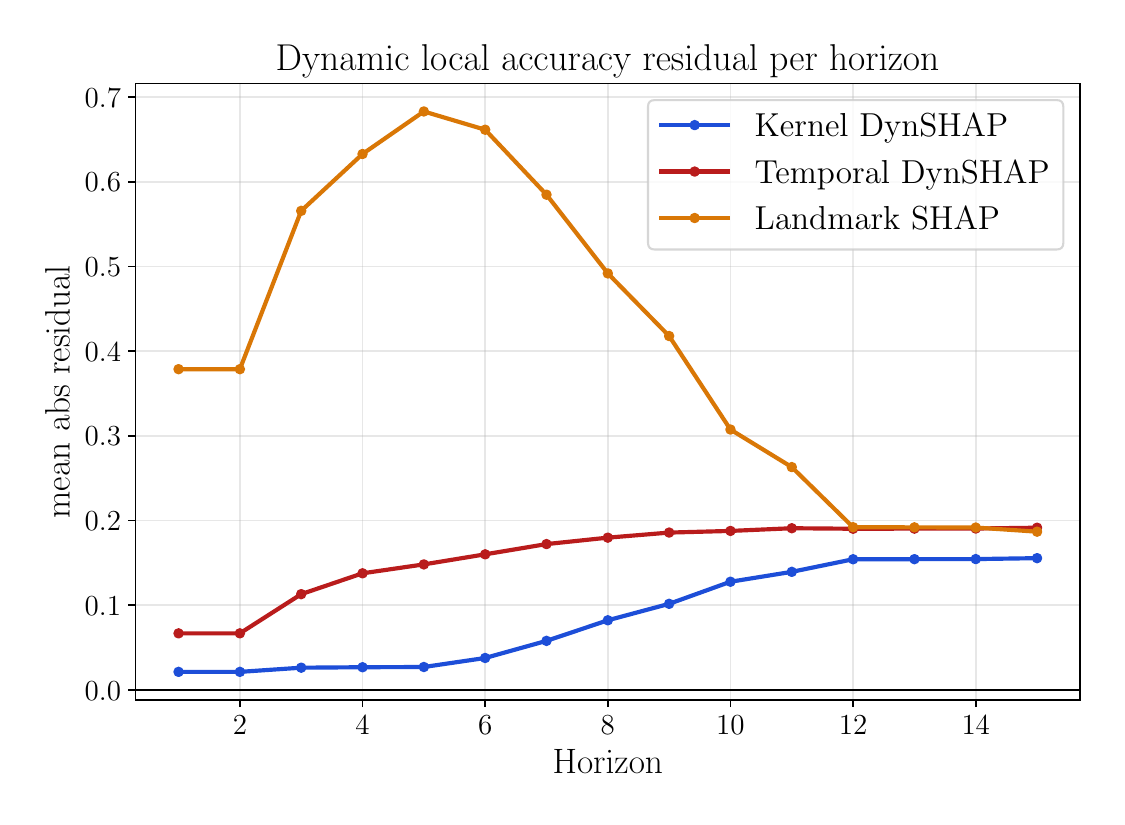}
        \caption{Local accuracy residuals, DySurv, PBC.}
        \label{fig:local_acc_pbc}
    \end{subfigure}
    \caption{\textbf{Comparison of local accuracy residuals ($\downarrow$) across horizons.} The x--axis shows the horizons at which the survival function is predicted.}
    \label{fig:local_acc_overall}
\end{figure}

Figure \ref{fig:local_acc_overall} shows the local accuracy residuals as a sanity check. It clearly demonstrates landmarking's violation of local accuracy. Local accuracy is desirable since it guarantees that there is no missing information and a fair credit distribution when combined with other SHAP properties. Formally, we define:

\begin{equation}
    \text{Mean abs residual} =\lvert f(\mathbf{x},\mathbf{t}_{out}) - \mathbb{E}[f(X,\mathbf{t}_{out})] - \sum_{i \in P} \phi_i(f,\mathbf{x})\rvert
\end{equation}

\section{Model Description}
\label{appendix-model-desc}

\subsubsection{Baseline Models}
\paragraph{Linear} XAI methods are mostly useful for the so-called `black box' models, where the logic behind model decisions remains ambiguous. However, to test the correctness of the implemented algorithms we require a simple model with known feature attributions to compare against. One such model we will use in tests is an additive linear model:
\begin{equation}
\label{eq:linear_model}
    f(x) = \sum_{t,i}\beta_{t,i} \times x_{t,i}
\end{equation}
for each $(t,i)$ time--feature pair.

The SHAP values for this model do not require approximation and are calculated using:
\begin{equation}
\label{eq:linear_shap}
    \phi_{t,i} = \beta_{t,i} \left (x_{t,i} - \mathbb{E}[X_{t,i}] \right )
\end{equation}

\subsubsection{Dynamic-DeepHit} 
\begin{figure} [htb]
    \centering
    \includegraphics[width=0.7\linewidth]{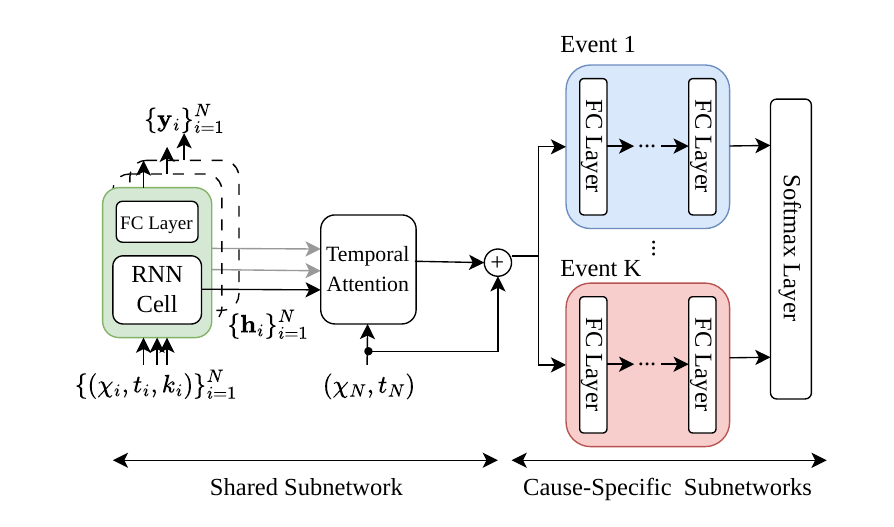}
    \caption{Network architecture of Dynamic-DeepHit. Adapted from Figure 2(a) \cite{DDH}.}
    \label{fig:ddh-architecture}
\end{figure}

\paragraph{Architecture} Dynamic-DeepHit \cite{DDH} is the first to investigate the application of deep learning to survival analysis with longitudinal measurements. This model adopts both RNNs and an attention mechanism to address the dynamic nature of the task, which makes it a representative  
example of a highly complex `black box' combining different methods to learn temporal dependencies. The detailed architecture of the model is outlined below:

\begin{enumerate}
  \item \textbf{Shared Subnetwork}: The shared subnetwork part of Dynamic-DeepHit's architecture consists of two components to process temporal dependencies: a RNN structure to tackle variable patients' history lengths and an attention mechanism to learn the longer-range dependencies.
  \item \textbf{Cause-Specific Subnetworks}: Each subnetwork is a fully-connected feed-forward network. These are designed to capture the complex relationships between a patient's historical measurements and the specific risk of a given clinical event. The $k$-th cause-specific subnetwork ingests the shared context vector $c$ alongside the final recorded measurement $x_J$ and its corresponding missingness indicator $m_J$. It processes these combined inputs to compute a cause-specific representation vector: $f_{c_k}(c, x_J, m_J)$.
  \item \textbf{Output Layer}: Finally, the model uses softmax to map the outcomes of each cause-specific network to proper probability measures.  
\end{enumerate}

\paragraph{Training} \label{loss} To train Dynamic-DeepHit, we minimise a total loss function $\mathcal{L}_{total}$ that is specifically designed to handle longitudinal measurements and right-censoring. It includes three terms: log-likelihood loss, ranking loss and prediction loss. The log-likelihood loss captures the joint distribution of the first hitting time and the corresponding event, simultaneously accounting for right-censored subjects. The ranking loss fine-tunes the network towards cause-specific discrimination by penalising incorrectly ordered pairs of subjects, encouraging higher predicted risks for those who experience an event sooner than others (compared at times elapsed since their last measurement). The prediction loss acts as an auxiliary regulariser on the shared subnetwork, requiring it to forecast step-ahead values of time-varying covariates so that the learned hidden representations preserve information about the longitudinal trajectories.

\subsubsection{DySurv}
\label{sec:dysurv}

\begin{figure}[htb]
    \centering
    \includegraphics[width=0.7\linewidth]{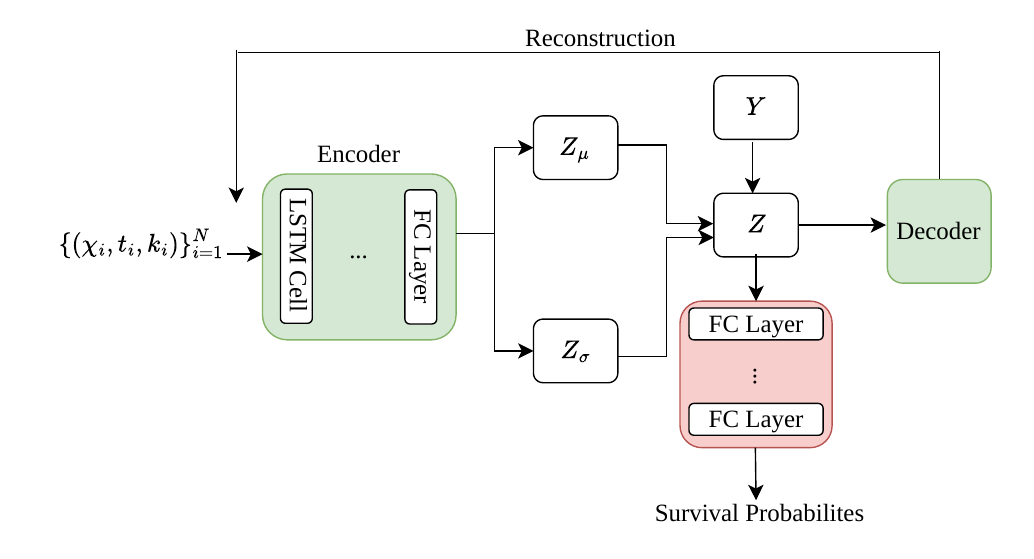}
    \caption{Network architecture of DySurv. Adapted from \url{https://github.com/munibmesinovic/DySurv/blob/main/DySurv.png}.}
    \label{fig:dysurv-architecture}
\end{figure}

\paragraph{Architecture} DySurv \cite{mesinovic2024dysurvdynamicdeeplearning} is a more recent model that combines RNNs (LSTM units) and a conditional variational autoencoder to handle temporal data. The architecture consists of:
\begin{enumerate}
  \item \textbf{LSTM Encoder}: Encodes the covariates in a hidden state.
  \item \textbf{Probabilistic Encoder}: Learns a Gaussian latent distribution parametrised by mean $\mu$ and standard deviation $\sigma$ and outputs a latent vector $Z$.
  \item \textbf{Multilayer Perceptron}: Predicts discrete-time cumulative incidence risk estimates $\hat{F}(t \mid \mathbf{x})$ across $K$ equally spaced time intervals.
\end{enumerate}

\paragraph{Training} \label{dysurv-loss} To train DySurv, we minimise a composite loss 
$\mathcal{L} = \alpha \mathcal{L}_{\text{surv}} + (1-\alpha)\mathcal{L}_{\text{VAE}}$, 
where $\alpha \in [0,1]$ is a tunable hyperparameter. The survival loss 
$\mathcal{L}_{\text{surv}}$ is a negative log-likelihood over the joint distribution of 
event time and outcome, accommodating right-censored observations without making any 
parametric or proportional hazards assumptions. The VAE loss $\mathcal{L}_{\text{VAE}}$ 
combines an MSE reconstruction term with a KL divergence penalty regularising the latent 
space toward a standard Gaussian.

\subsection{Training Configuration}

We perform a search over common hyperparameters for the two `black box' models: Dynamic-DeepHit and DySurv. The aim is to achieve a high time-dependent concordance index ($\geq$ 0.7) to announce the models accurate and interesting candidates for explanations. Table~\ref{tab:ddh-hyperparams} and Table~\ref{tab:dysurv-hyperparams} outline the final hyperparameters selected for both models for each dataset.

\begin{table}[h]
    \centering
    \begin{tabular}{l|lll}
    \toprule
    Hyperparameter & PBC & MS & Synthetic \\
    \midrule
         Num. RNN layers & 1 & 4 & 2 \\
         RNN hidden size & 50 & 32 & 50 \\
         Dropout & 0.16 & 0.16 & 0.16 \\
         Optimiser & Adam & Adam & Adam \\
         Learning rate & 2e-3 & 1e-3 & 1e-3 \\
         Batch size & 512 & 32 & 32 \\
         Epochs & 50 & 50 & 200 \\
    \bottomrule
    \end{tabular}
    \vspace{6pt}
    \caption{Hyperparameters used to train Dynamic-DeepHit on each dataset.}
    \label{tab:ddh-hyperparams}
\end{table}

\begin{table}[h]
    \centering
    \begin{tabular}{l|lll}
    \toprule
    Hyperparameter & PBC & MS & Synthetic \\
    \midrule
         $\alpha_{\text{surv}}$ & 1.0 & 1.0 & 1.0 \\
         $\alpha_{\text{ae}}$ & 0.5 & 0.5 & 0.5 \\
         $\alpha_{\text{kl}}$ & 0.16 & 0.001 & 0.16 \\
         Optimiser & Adam & Adam & Adam \\
         Learning rate & 1e-3 & 1e-3 & 1e-3 \\
         Batch size & 64 & 64 & 64 \\
         Epochs & 200 & 200 & 200 \\
    \bottomrule
    \end{tabular}
    \vspace{6pt}
    \caption{Hyperparameters used to train DySurv on each dataset.}
    \label{tab:dysurv-hyperparams}
\end{table}

\subsection{Training Results}
\label{appendix-model-training}

The corresponding metrics for every model trained on each dataset are summarised in Table \ref{tab:training-results}.

\begin{table}[htb]
\centering
\small
\begin{tabular}{l c c c c}
\toprule
& \multicolumn{2}{c}{\textbf{Dynamic-DeepHit}} & \multicolumn{2}{c}{\textbf{DySurv}} \\
\cmidrule(lr){2-3} \cmidrule(lr){4-5}

\textbf{Dataset} & C-Index ($\uparrow$) & IBS ($\downarrow$) & C-Index ($\uparrow$) & IBS ($\downarrow$) \\
\midrule
PBC & $\mathbf{0.9419 \pm 0.0069}$ & $\mathbf{0.0637 \pm 0.0029}$ & $0.8491 \pm 0.0280$ & $0.1595 \pm 0.0126$ \\
MS & $0.7656 \pm 0.0082$ & $0.3356 \pm 0.0220$ & $\mathbf{0.7823 \pm 0.0265}$ & $\mathbf{0.3120 \pm 0.1549}$\\
Synthetic Data & $\mathbf{0.7957 \pm 0.0024}$& $\mathbf{0.1397 \pm 0.0016}$& $0.7125 \pm 0.0216$ & $0.2581 \pm 0.0058$\\
\bottomrule
\end{tabular}
\vspace{6pt}
\caption{\textbf{Aggregated time-dependent performance of Dynamic-DeepHit and DySurv trained on all datasets.} Measured by time-dependent concordance index and integrated Brier score. Reported mean $\pm$ one standard deviation.}
\label{tab:training-results}
\end{table}

\section{Synthetic Data and Ground Truth Definition}
\label{synthetic-data-generation}

Real-world datasets and `black box' models have no established ground truth against which we can compare explanations. Therefore, we require the use of synthetic data with defined ground truth. This section details the synthetic data design and generation. 

\subsubsection{Synthetic Data Generation}
\label{sec:synth-data}
Ideally, the synthetic data should be clinically-accurate and should include features which match the underlying assumption of Temporal DynSHAP. We generate a synthetic longitudinal survival dataset of $N=2000$ patients with irregular visit schedules, following established methods for simulating plausible right-censored clinical data with time-varying covariates \cite{crowther2013simulating}. The dataset is designed with four covariates, each targeting a specific aspect of the SHAP estimators: a temporally autocorrelated continuous feature to test Temporal DynSHAP, a binary treatment indicator with a time-dependent onset to test temporal attribution, a static risk factor, and a noise feature with zero true effect.

The hazard function for each patient follows a Weibull proportional hazards model:
\begin{equation}
\label{eq:synth_hazard}
    h(t) = \underbrace{\lambda \gamma \, t^{\gamma - 1}}_{h_0(t)} \cdot \exp\!\Big(\beta_0 \, F_0(t) + \beta_1 \, F_1(t) + \beta_2 \, F_2 + \beta_3 \, F_3\Big),
\end{equation}
where $h_0(t) = \lambda \gamma \, t^{\gamma-1}$ is the Weibull baseline hazard. The four covariates are defined as follows:

\paragraph{$F_0$: Autocorrelated biomarker ($\beta_0 = 2.5$).} A continuous time-varying covariate generated by a continuous-time AR(1) process:
\begin{align}
    F_0(t_0) &\sim \mathcal{N}(0, 1), \nonumber \\
    F_0(t_i) &= \alpha^{\Delta t} \, F_0(t_{i-1}) + \sqrt{1 - \alpha^{2\Delta t}} \cdot \varepsilon, \quad \varepsilon \sim \mathcal{N}(0, \sigma^2),
    \label{eq:ar1}
\end{align}
where $\Delta t = t_i - t_{i-1}$ and $\alpha = 0.85$ for strong autocorrelation, which creates strong temporal dependence between consecutive observations. Marginal SHAP estimators, which sample replacement values independently across time, produce out-of-distribution counterfactuals for this feature, while Temporal DynSHAP should be able to preserve the temporal structure.

\paragraph{$F_1$: Treatment indicator ($\beta_1 = -2.5$).} A binary covariate that switches from 0 to 1 at a patient-specific time $t_s$, representing treatment onset:
\begin{equation}
    F_1(t) = \mathbf{1}[t \geq t_s], \quad \text{where } t_s \sim \mathrm{Uniform}(1, 6) \text{ or } t_s = \infty \text{ (untreated)}.
\end{equation}
The negative coefficient means treatment decreases risk. This feature tests whether the estimators correctly localise attribution to the time window after treatment activation.

\paragraph{$F_2$: Static risk factor ($\beta_2 = 0.5$).} A time-invariant binary covariate drawn at baseline. Its SHAP attribution should be constant across time bins.

\paragraph{$F_3$: Noise ($\beta_3 = 0$).} A time-invariant draw from $\mathcal{N}(0,1)$ with no effect on the hazard. Any non-zero SHAP attribution to this feature indicates shortcut learning.

\medskip

Event times are generated via the inverse cumulative hazard method \cite{event_times_bender}: we draw $U \sim \mathrm{Uniform}(0,1)$, set the target cumulative hazard to $-\log U$, and solve $H(T) = -\log U$ by bisection, where $H(T) = \int_0^T h(s)\,ds$ is computed numerically with $F_0(t)$ linearly interpolated between visits. Administrative right-censoring is applied to 30\% of patients. Visit times are continuous timestamps with inter-visit gaps drawn from $\mathrm{Uniform}(0.5, 2.0)$.

\subsubsection{Ground Truth SHAP}
\label{sec:gt-shap}

We can now define the ground truth SHAP values against which we can benchmark the estimates in the future. We use the conditional value function to do this. The true probability distributions of all features are defined in Section~\ref{sec:synth-data}. This allows us to use the exact conditional value function instead of a learned approximation (what Temporal DynSHAP does). Specifically, for each coalition mask, if the time--feature pair is in coalition (its value in the coalition mask is 1), we keep the test patient's value. If the time--feature pair is not in coalition (indicated by mask value 0) we sample from the true distributions we used to generate the data. This is exactly the same conditional sampling process we use in the implementation of Temporal DynSHAP except the sampling is done from the true data distribution instead of a learned approximation. The true SHAP values are then approximated using Equation~\ref{eq:perm-dyn}.

\section{Exploring Temporal Dependence in Data}
\label{appendix-temp-dep}
This section presents the results of exposing linear temporal dependencies in the real-world medical data used for evaluation in this paper. These emphasise the unsuitability of marginal SHAP to time series explanations and indicate the need for estimators aware of temporal dependencies, such as Temporal DynSHAP presented. 

\begin{figure}
     \centering
     \begin{subfigure}[t]{0.49\textwidth}
         \centering
         \includegraphics[width=\textwidth]{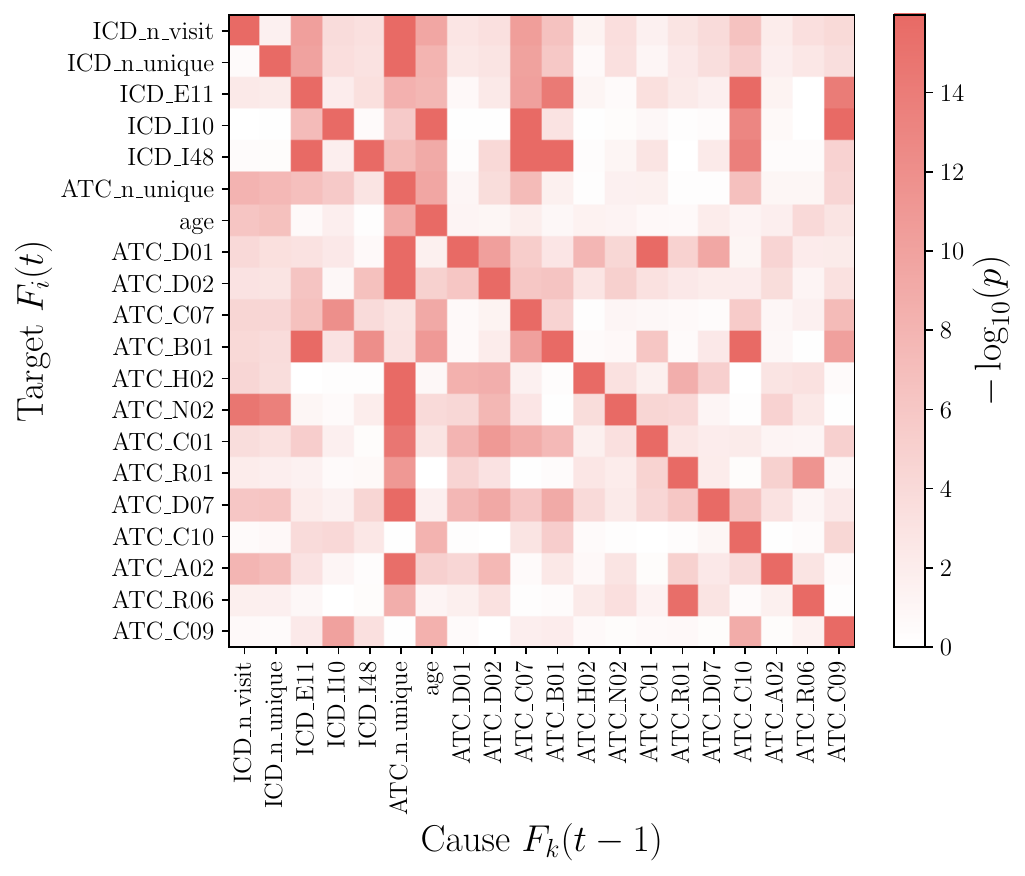}
         \caption{Granger test results for the MS dataset.}
         \label{fig:granger_ms}
     \end{subfigure}
     \hfill
     \begin{subfigure}[t]{0.49\textwidth}
         \centering
         \includegraphics[width=\textwidth]{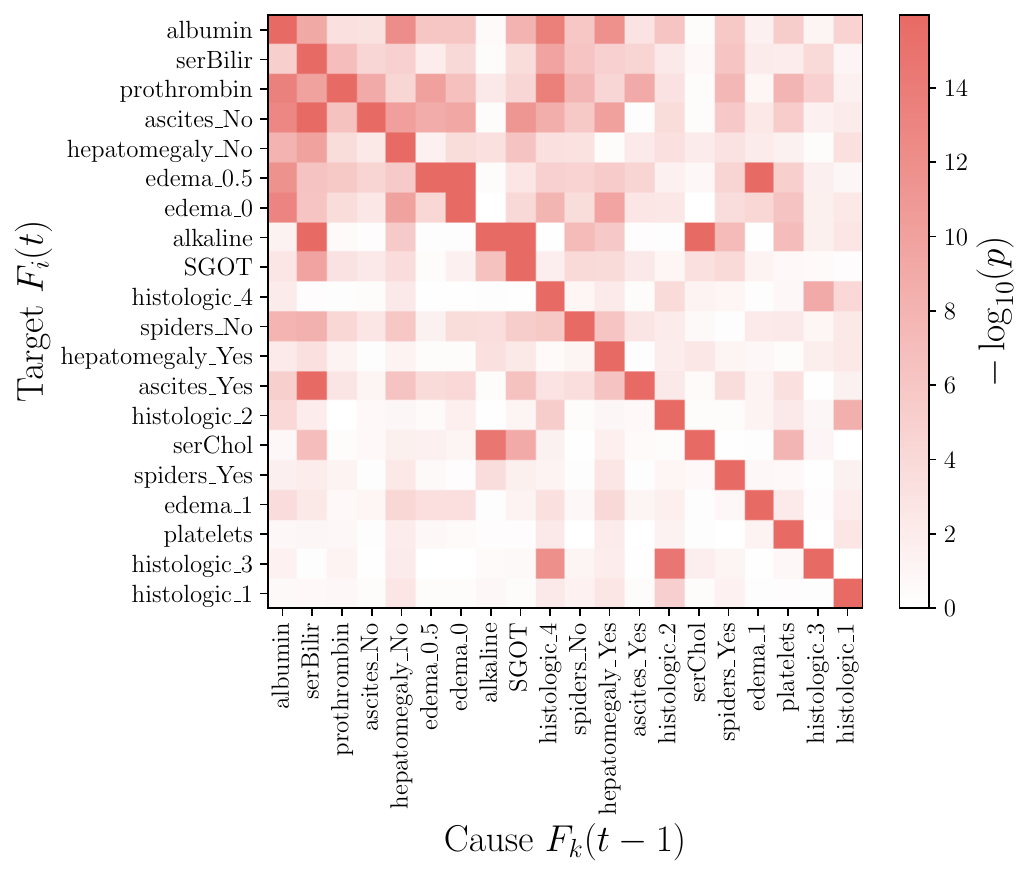}
         \caption{Granger test results for the PBC dataset.}
         \label{fig:granger_pbc}
     \end{subfigure}
    \caption{Pairwise Granger test results sorted for the top 20 features by significance. Each cell (j,k) shows $-\log_{10}(p)$ for the Null hypothesis that lagged feature $X_k(t-1)$ does not predict target feature $X_j(t)$. Darker cells indicate stronger evidence of temporal predictive dependence.}
    \label{fig:temporal-dep-pbc-ms}
\end{figure}

In particular, we explore the feature dependence in the time domain by running Granger causality tests \cite{granger} for each model feature. By isolating consecutive medical visits (time $t-1$ and time $t$) for each patient and pooling these temporal pairs across the entire cohort, we perform an F-test to determine if a feature's historical value significantly improved the prediction of its current value. Formally:
\begin{equation}
    F_i(t) = A_kF_k(t-1)
\end{equation}
$X_k(t-1)$ is then said to Granger-cause $F_i(t)$ if $A_k$ is significantly larger than zero in absolute value, where ``significance'' is decided by the hypothesis test. The Null hypothesis is that $F_k(t-1)$ does not Granger-cause $F_i(t)$. We reject the Null hypothesis if the $p$-value is lower than $0.05$. We use the \texttt{grangercausalitytests} library to run the tests. The plots in Figure~\ref{fig:temporal-dep-pbc-ms} demonstrate the results for the two real-world datasets.

Granger causality does not equal true causality \cite{granger-fallacy}. However, this test indicates whether the past features can be used to predict the future ones, which suggests temporal dependence. The results demonstrate that features are indeed not fully independent across time, motivating a new temporal estimator.

\section{Additional Evaluation of DynSHAP}

This section presents additional results for DynSHAP evaluation. All experiments in the paper are run across 5 seeds: $\{42, 1, 2, 3, 4\}$.

\subsection{DynSHAP Correctness}
\label{appendix-correctness}
We perform sanity checks to ensure the estimators are implemented correctly. Table \ref{tab:linear-model-res} demonstrates that the two marginal estimators converge to exact closed-form Shapley values of a linear additive model values with negligible error. Temporal DynSHAP, on the other hand, produces a larger RMSE of 0.0498. One explanation could be that the fitted linear regressions used for conditional sampling create random false dependencies where there are none.

\begin{table}[htb]
\centering
\begin{tabular}{l|c}
\toprule
    Estimator & RMSE ($\downarrow$) \\
\midrule
    Sampling, $\times10^{-15}$ & $\mathbf{0.0475 \pm 0.0083}$ \\
    Kernel, $\times10^{-13}$ & 0.0837 $\pm 0.0514$\\
    Temporal & 0.0498 $\pm 0.0236$ \\
\end{tabular}
\vspace{6pt}
\caption{\textbf{Comparison between SHAP estimates and ground truth SHAP values.} Evaluated on a linear additive model with closed-form Shapley values trained on independent time--feature data drawn from $\mathcal{N}[0,1]$}
\label{tab:linear-model-res}
\end{table}

\subsection{Robustness to Noise}
\label{appendix-ris}

\begin{figure}[htbp]
     \centering
     \begin{subfigure}[t]{0.3\textwidth}
         \centering
         \includegraphics[width=\textwidth]{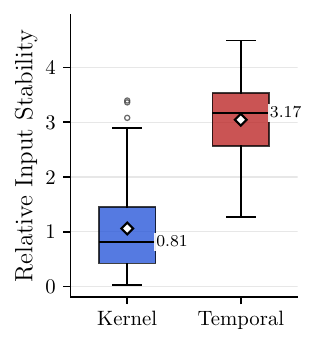}
         \caption{Log RIS, synthetic dataset}
         \label{fig:synth-ris}
     \end{subfigure}
     \hfill
     \begin{subfigure}[t]{0.3\textwidth}
         \centering
         \includegraphics[width=\textwidth]{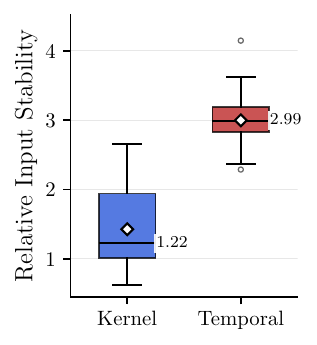}
         \caption{Log RIS, PBC dataset}
         \label{fig:pbc-ris}
     \end{subfigure}
     \hfill
     \begin{subfigure}[t]{0.3\textwidth}
         \centering
         \includegraphics[width=\textwidth]{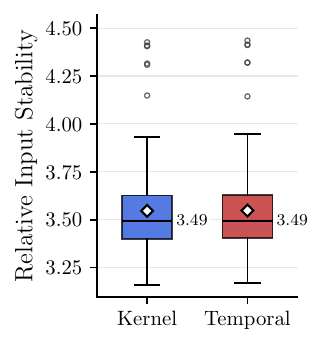}
         \caption{Log RIS, MS dataset}
         \label{fig:ms-ris}
     \end{subfigure}
     \caption{\textbf{Log Relative Input Stability (RIS) $(\downarrow)$.} Calculated for the marginal estimator (Kernel DynSHAP) and the conditional one (Temporal DynSHAP) by perturbing inputs by adding Gaussian noise.}
     \label{fig:ris-plots}
\end{figure}

We further employ the Relative Input Stability (RIS) metric \citep{agarwal2022rethinkingstabilityattributionbasedexplanations} to measure the tool's stability, results summarised in Figure \ref{fig:ris-plots}. For the synthetic and PBC datasets, Temporal DynSHAP produces less stable attributions than Kernel DynSHAP, with median log RIS roughly 3$\times$ higher. Temporal DynSHAP relies on the fitted linear regressions to impute out-of-coalition values, which could explain a drop in stability, since the regressions' fitted parameters introduce an additional source of variance. On the MS dataset, both methods produce comparable, less stable explanations (log RIS  3.49). This might be driven by the dataset's sparsity, which could lead to model overfitting and the prediction drastically changing in response to any input perturbation, simultaneously changing the SHAP estimates. The range in which the log RIS results lie for both estimators is considered stable in literature \cite{agarwal2024openxaitransparentevaluationmodel}.

\subsection{Computational efficiency}
\label{appendix-comp-eff}
\begin{figure}[htb]
    \centering
    \includegraphics[width=\linewidth]{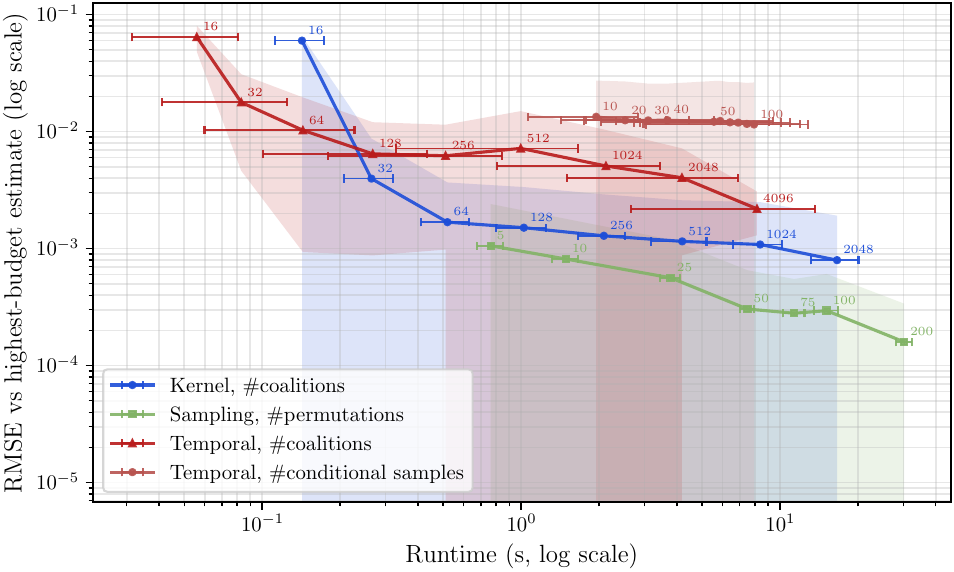}
    \caption{\textbf{Convergence analysis of the dynamic SHAP estimators over parameters used for computational speedup.} Measured with RMSE ($\downarrow$) for DDH trained on the synthetic data. Run across 5 different seeds, reported mean $\pm$ std for both runtime and RMSE.}
    \label{fig:convergence}
\end{figure}

\begin{figure}[htb]
    \centering
    \includegraphics[width=\linewidth]{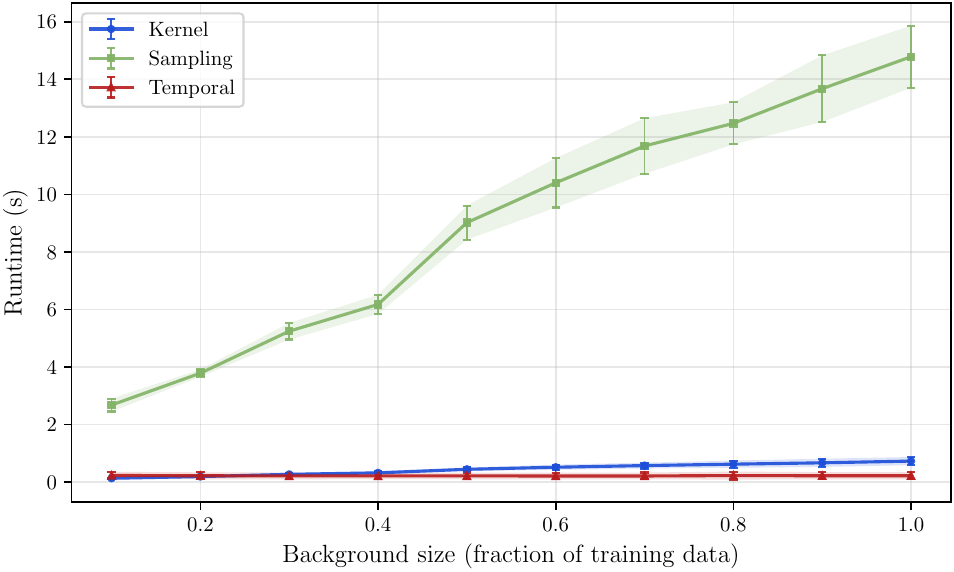}
    \caption{\textbf{Runtime analysis over the background size parameter used for computational speedup.} Measured for DDH trained on the synthetic data over five different seeds.}
    \label{fig:conv-runtime}
\end{figure}

A significant limitation of SHAP is its computational inefficiency. In addition to addressing temporal feature dependence, Temporal DynSHAP's sampling process reduces runtime, as presented in Figures \ref{fig:convergence} and \ref{fig:conv-runtime}.

\subsection{MS Case Study}
\label{appendix-ms-case-study}

\begin{figure}[htbp]
    \centering
    \begin{subfigure}[b]{0.30\textwidth}
        \centering
        \includegraphics[width=\linewidth]{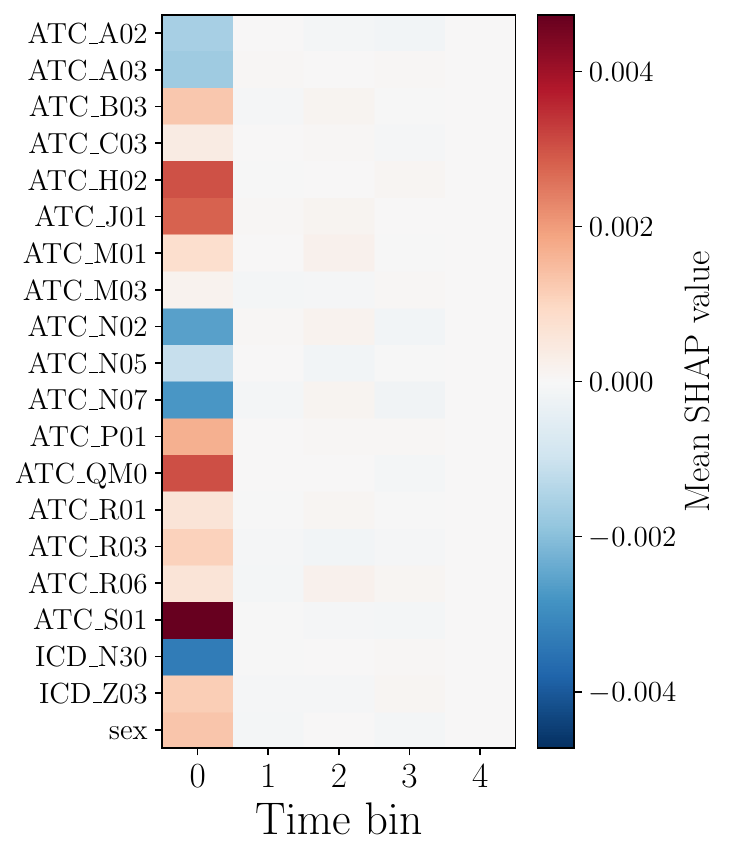}
        \caption{Kernel DynSHAP, DDH}
        \label{fig:kernel-ddh-85}
    \end{subfigure}
    \hfill
    \begin{subfigure}[b]{0.30\textwidth}
        \centering
        \includegraphics[width=\linewidth]{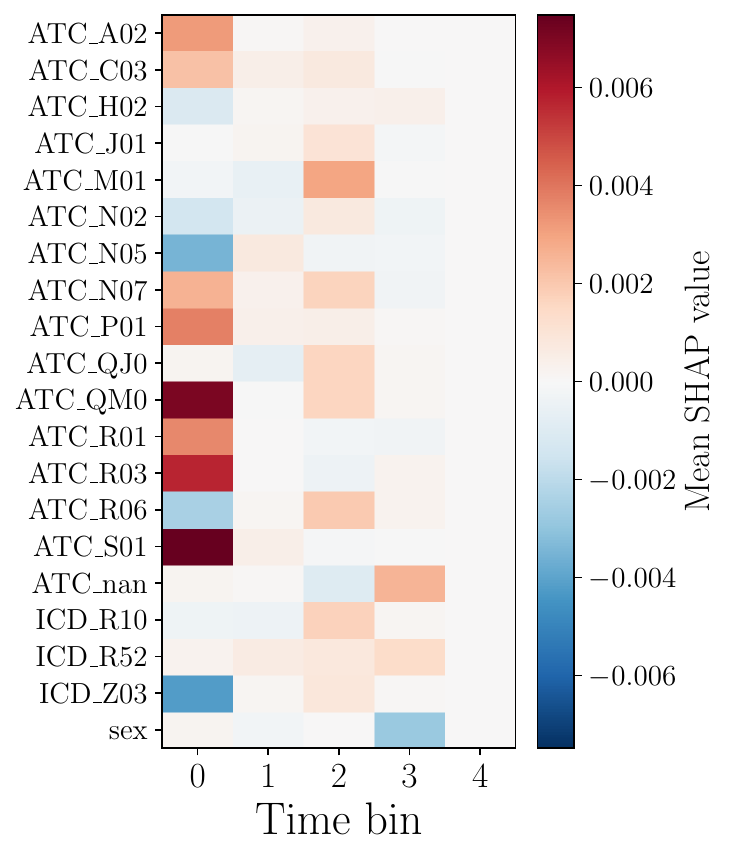}
        \caption{Temporal DynSHAP, DDH}
        \label{fig:temporal-ddh-85}
    \end{subfigure}
    \hfill
         \begin{subfigure}[b]{0.30\textwidth}
        \centering
        \includegraphics[width=\linewidth]{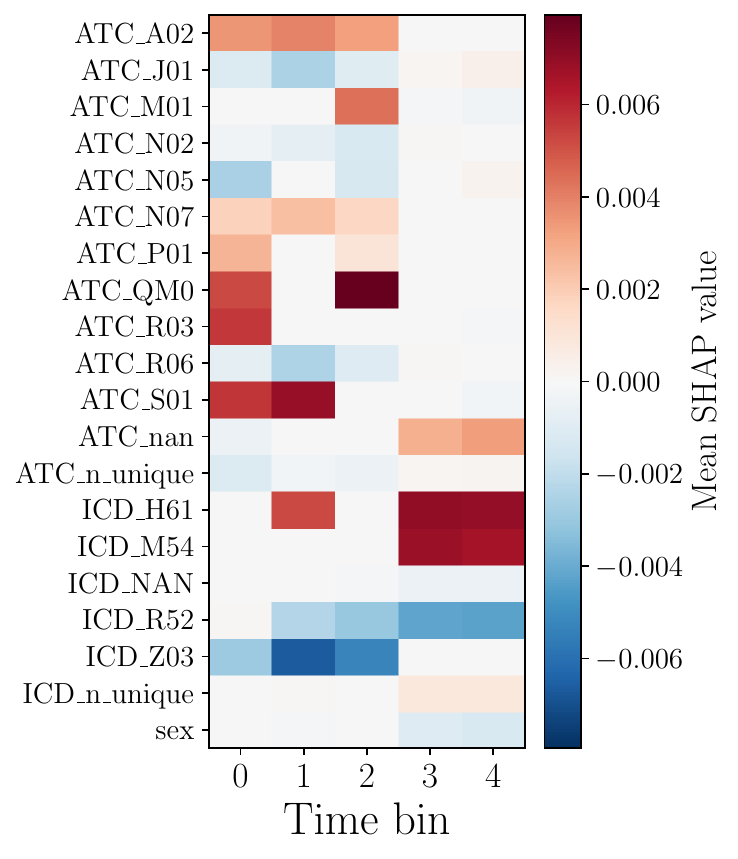}
        \caption{Landmark SHAP, DDH}
        \label{fig:lm-ddh-85}
    \end{subfigure}
    
    \vspace{1em} 

    \begin{subfigure}[b]{0.30\textwidth}
        \centering
        \includegraphics[width=\linewidth]{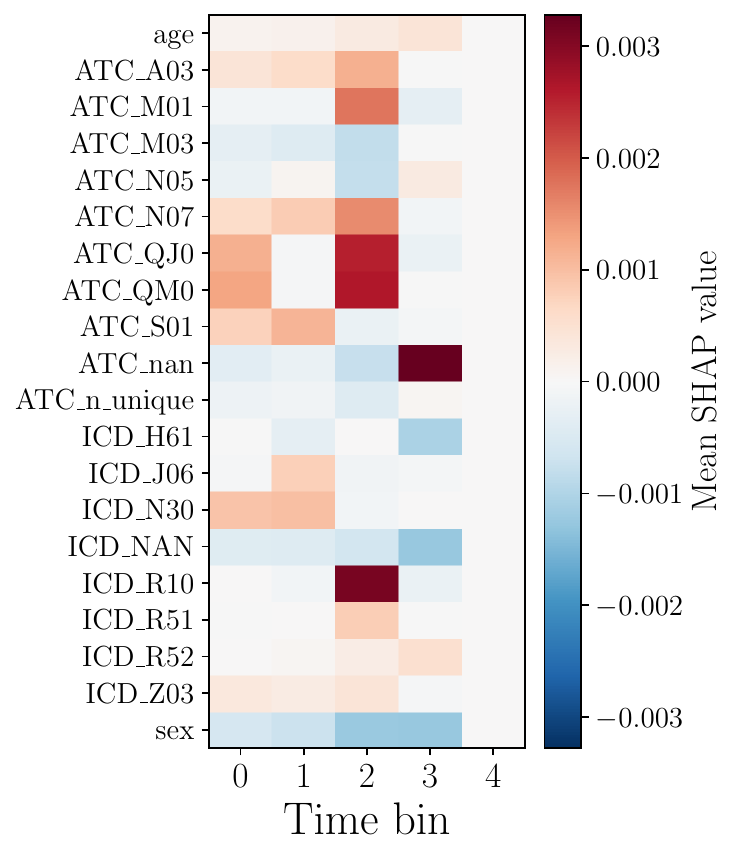}
        \caption{Kernel DynSHAP, DySurv}
        \label{fig:kernel-ds-85}
    \end{subfigure}
    \hfill
    \begin{subfigure}[b]{0.30\textwidth}
        \centering
        \includegraphics[width=\linewidth]{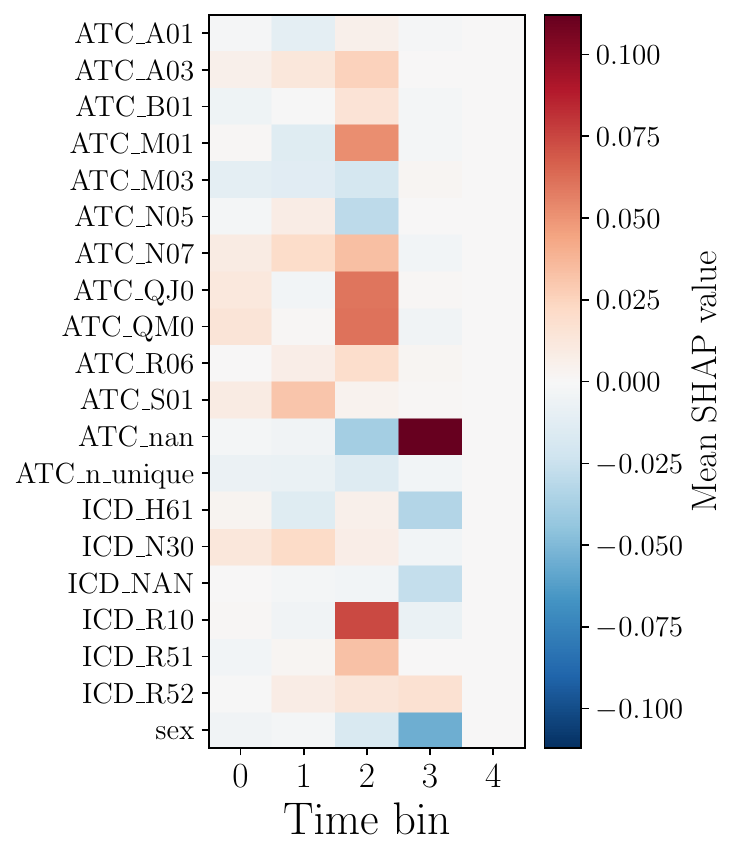}
        \caption{Temporal DynSHAP, DySurv}
        \label{fig:temporal-ds-85}
    \end{subfigure}
    \hfill
         \begin{subfigure}[b]{0.30\textwidth}
        \centering
        \includegraphics[width=\linewidth]{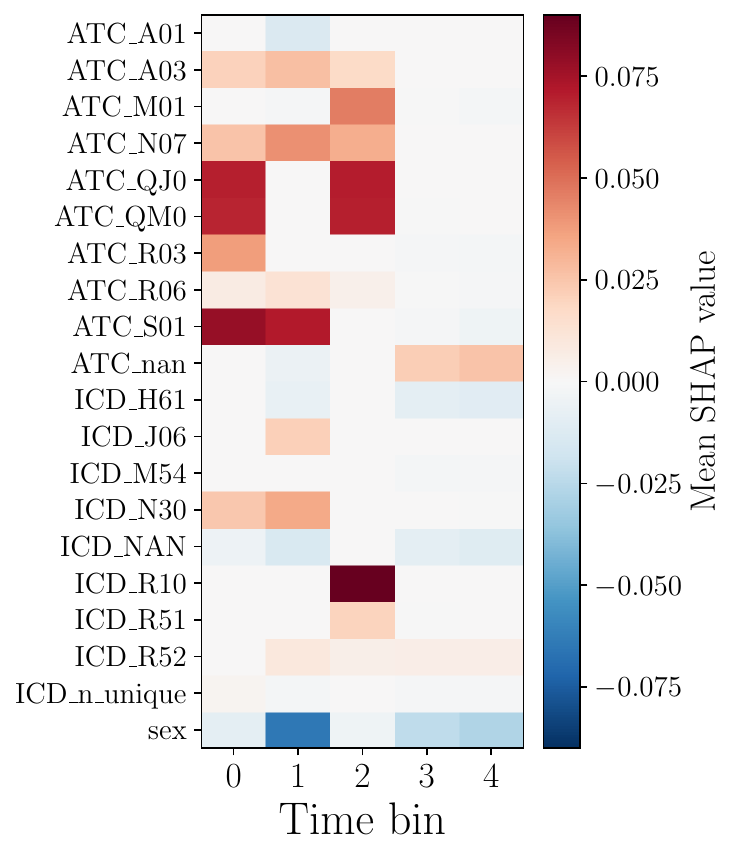}
        \caption{Landmark SHAP, DySurv}
        \label{fig:lm-ds-85}
    \end{subfigure}

    \caption{\textbf{A comparison of the estimators applied to one MS-positive patient.}}
    \label{fig:case-study}
\end{figure}

Figure \ref{fig:case-study} presents the DynSHAP estimations produced by the Kernel and Temporal frameworks applied to DDH and DySurv for an MS-positive patient alongside the landmarking approach. As previously verified, DynSHAP's explanations reveal model history learning better, so one can analyse them to make judgements about architectures. For example, Figure \ref{fig:case-study} reveals that DySurv uses more history in making its prediction than DDH. We also note that the features with top attributions are clinically valid. For example, certain medications in the ATC R03 category are associated with a reduced risk of MS \cite{drugms} and both estimators identify them as key factors for survival in DDH. There are also some  examples of DDH exhibiting a spurious correlation. For example, ICD Z03 (Medical observation and evaluation) is associated with reduced survival (higher MS risk). It is not a cause of MS but rather simply indicates that the patients seek medical help more, which we naturally observe in patients that are likely to develop a disease. 

\newpage

\end{document}